\documentclass[10pt,twocolumn,letterpaper]{article}

\newif\ifproceedings  %
\newif\ifaddappendix

\addappendixtrue
\proceedingsfalse

\ifproceedings
    \usepackage{cvpr}              %
\else
    \usepackage[pagenumbers]{cvpr} %
\fi

\usepackage{graphicx}
\usepackage{booktabs}  %
\usepackage[accsupp]{axessibility}  %
\usepackage[utf8]{inputenc} %
\usepackage{amsfonts}       %
\usepackage{nicefrac}       %
\usepackage{microtype}      %
\usepackage{gensymb}
\usepackage{tabularx}
\usepackage{multirow}
\usepackage{svg}
\usepackage{cuted}
\usepackage{tabu}
\usepackage{pifont}  %
\usepackage{mwe}
\usepackage{bm}  %
\usepackage{textcomp}
\usepackage{makecell}  %
\usepackage{grffile} %
\usepackage{rotating}  %
\usepackage{dsfont}  %
\usepackage[normalem]{ulem} %
\usepackage{float}
\usepackage{colortbl}

\usepackage{siunitx}  %
\newif\ifshowcomments
\showcommentstrue 

\newcommand{\comment}[1]{}

\renewcommand{\*}[1]{\bm{\mathrm{#1}}}
\renewcommand{\b}[1]{\textbf{#1}}
\newcommand{\red}[1]{\textcolor{red}{#1}}

\newcommand{\blue}[1]{\textcolor{blue}{#1}}
\newcommand{\0}{\phantom{0}}
\newcommand{\mytilde}{{\raise.17ex\hbox{$\scriptstyle\sim$}}}

\renewcommand{\paragraph}[1]{\vskip4pt \noindent\textbf{#1}}

\definecolor{tabfirst}{rgb}{1, 0.7, 0.7} %
\definecolor{tabsecond}{rgb}{1, 0.85, 0.7} %
\definecolor{tabthird}{rgb}{1, 1, 0.7} %

\newlength{\pheight}
\newlength{\pwidth}
\newlength{\lwidth}
\newlength{\bwidth}
\newlength{\iwidth}

\newcommand{\real}{\mathbb{R}}

\newcommand{\norm}[1]{\left\lVert#1\right\rVert}

\newcommand{\set}[1]{\{ #1 \}} %

\newcommand{\registered}{\mathcal{R}}

\definecolor{cvprblue}{rgb}{0.21,0.49,0.74}
\usepackage[pagebackref,breaklinks,colorlinks,allcolors=cvprblue,hyperfootnotes=false]{hyperref}

\def\paperID{10202} %
\def\confName{CVPR}
\def\confYear{2025}

\title{MP-SfM: Monocular Surface Priors for Robust Structure-from-Motion}

\author{%
Zador Pataki$^{1}$\hspace{0.3in}%
Paul-Edouard Sarlin$^{2}$\hspace{.3in}%
Johannes L.~Sch\"onberger$^{1,3}$\hspace{.3in}%
Marc Pollefeys$^{1,3}$
\vspace{0.05in}\\
$^{1}$ ETH Zurich\hspace{0.3in}
$^{2}$ Google\hspace{0.3in}
$^{3}$ Microsoft Spatial AI Lab
}

\begin{document}
\maketitle

\begin{abstract}

While Structure-from-Motion (SfM) has seen much progress over the years, state-of-the-art systems are prone to failure when facing extreme viewpoint changes in low-overlap, low-parallax or high-symmetry scenarios.
Because capturing images that avoid these pitfalls is challenging, this severely limits the wider use of SfM, especially by non-expert users.
We overcome these limitations by augmenting the classical SfM paradigm with monocular depth and normal priors inferred by deep neural networks.
Thanks to a tight integration of monocular and multi-view constraints, our approach significantly outperforms existing ones under extreme viewpoint changes, while maintaining strong performance in standard conditions.
We also show that monocular priors can help reject faulty associations due to symmetries, which is a long-standing problem for SfM.
This makes our approach the first capable of reliably reconstructing challenging indoor environments from few images.
Through principled uncertainty propagation, it is robust to errors in the priors, can handle priors inferred by different models with little tuning, and will thus easily benefit from future progress in monocular depth and normal estimation.
Our code is publicly available at~\href{https://github.com/cvg/mpsfm}
{\nolinkurl{github.com/cvg/mpsfm}}.

\end{abstract}

\vspace{-10pt}

\section{Introduction}

Structure-from-Motion (SfM) is a prevalent problem in computer vision involving the estimation of 3D structure and camera motion from a collection of 2D images.
The tremendous progress in the field has culminated in a variety of state-of-the-art SfM pipelines (\eg, Bundler~\cite{snavely2006photo}, VisualSfM~\cite{wu2011visualsfm}, COLMAP~\cite{schoenberger2016sfm}, GLOMAP~\cite{pan2024glomap}).
Today, these systems are successfully applied in a wide range of scenarios with high relevance for tasks such as (simultaneous) localization and mapping~\cite{sarlin2022lamar}, multi-view stereo~\cite{schoenberger2016mvs}, or novel-view synthesis~\cite{mildenhall2020nerf,kerbl3Dgaussians}.
Despite this feat, many challenges and failure cases remain to be solved, including but not limited to extreme viewpoint and illumination changes~\cite{superpoint,dusmanu2019d2,sarlin2020superglue}, repetitive structure~\cite{zach2010disambiguating,cai2023doppelgangers}, scalability to large scenes~\cite{agarwal2011building,frahm2010building,heinly2015reconstructing}, or privacy concerns~\cite{speciale2019privacy,geppert2020privacy}.

One of the most frequent failure cases for SfM is the scenario of extreme viewpoint changes.
These can manifest as either extremely wide-baseline and low-overlap or as low-parallax image pairs.
As such, these scenarios present challenges to different stages in the reconstruction process.
Many recent efforts in the community have centered around solving the matching problem using the immense progress in machine learning~\cite{superpoint,dusmanu2019d2,sarlin2020superglue,dust3r_cvpr24}, and we are now able to match across extreme viewpoint and illumination conditions or even detect symmetry issues from a single pair of images~\cite{cai2023doppelgangers}.
However, when we feed the established matches into the subsequent reconstruction algorithms, we still face multiple fundamental limitations leading to unstable and inaccurate results or even outright reconstruction failures.

\begin{figure}[t]
    \centering
    \includegraphics[width=\linewidth]{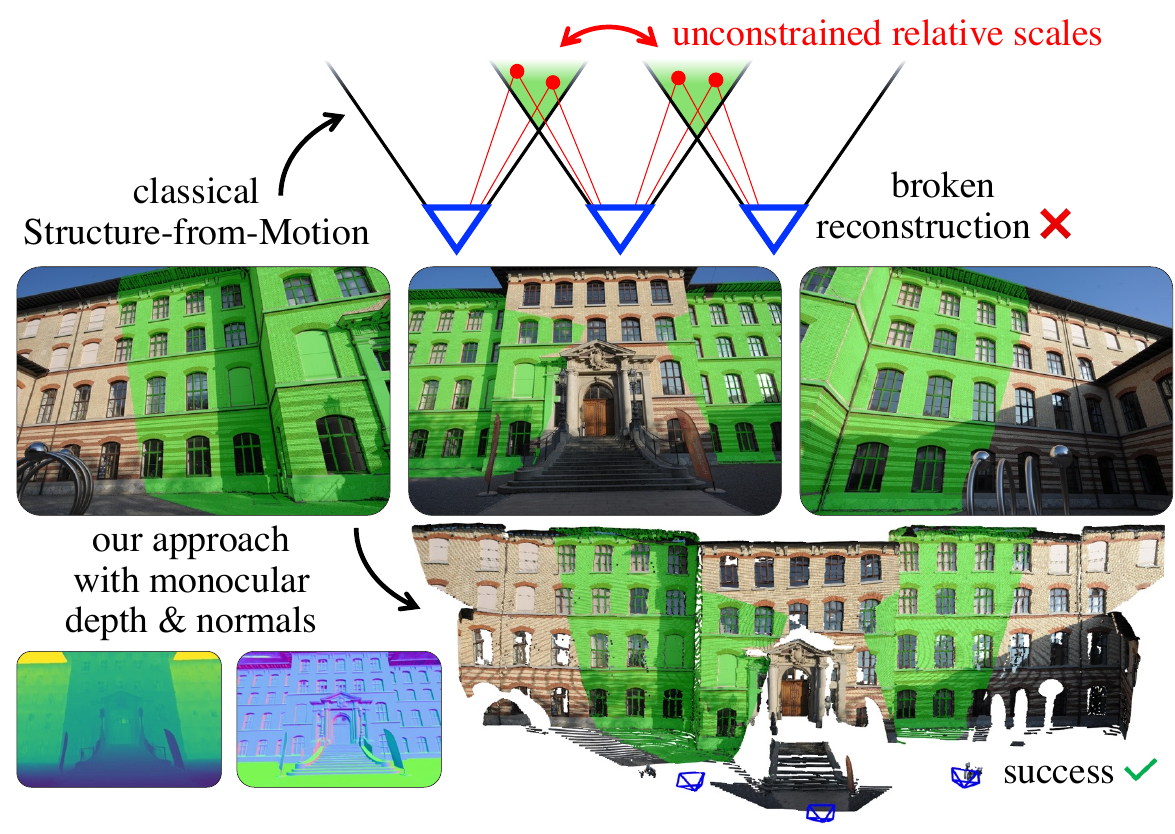} 
    \caption{\textbf{A typical failure case for SfM.}
    Existing approaches cannot handle low-overlap image pairs because they require three-view tracks to ensure a consistent scale across the scene.
    We bridge this limitation by augmenting SfM with monocular depth and normal priors from off-the-shelf deep networks.
    This makes SfM significantly more robust for data captured by non-expert users.
    }%
    \label{fig:teaser}%
\end{figure}

{In particular, the current state-of-the-art systems~\cite{schoenberger2016sfm,pan2024glomap,duisterhof2024mast3r,wang2023vggsfm} inherently require three-view overlap with sufficient baseline and parallax or otherwise cannot perform multi-view consistent 3D reconstruction.}
In practice, this turns out as one of the main difficulties for the non-expert user and oftentimes also for experts for a variety of reasons.
First, it is intrinsically hard to capture large and complex scenes while ensuring sufficient viewpoint overlap and variation.
Even for small scenes and structured setups, it is far from trivial with the many constraints to satisfy and it often requires careful prior planning or repeated trials to capture a scene with the desired completeness and accuracy.
The naive approach of capturing overly redundant viewpoints is typically also not a solution, as it stands in opposition to other important considerations like capture and processing time or storage and compute costs.
Furthermore, the general purpose reconstruction systems frequently show diminishing returns when fed with too many redundant views.

In this paper, we make an important step towards overcoming several of the remaining limitations by leveraging the recent advances in monocular depth estimation.
In particular, we integrate monocular depth and normal cues into the classical incremental SfM paradigm to lift the requirement for three-view tracks.
Our proposed pipeline is able to perform accurate multi-view 3D reconstruction from two-view tracks only and thus works in extremely challenging low-overlap scenarios, while retaining state-of-the-art performance in higher-overlap conditions (\cref{fig:teaser}).
As a consequence, our system can also directly take advantage of dense pairwise matches~\cite{sun2021loftr,dust3r_cvpr24,edstedt2024roma}.
Compared to the classical approach of only using sparse feature matches, we thus achieve higher reconstruction completeness in scenes with little texture.
In addition, the use of single-view depth priors for regularization of scene geometry leads to significant reliability improvements under low-parallax conditions.
We further introduce a dense depth consistency check to identify incorrect posing of images, especially to prevent symmetry issues.

By tight integration of single- and multi-view optimization and principled uncertainty propagation, we handle large errors in the monocular priors, which in turn enables us to build upon off-the-shelf deep models.
Future progress in monocular depth estimation will benefit our approach with little to no tuning required.
To maintain the scalability characteristics of classical SfM, we formulate the global reconstruction objective as an alternating optimization of single- and multi-view sub-problems.
In extensive experiments on challenging datasets, we show significant improvements in terms of accuracy and completeness as compared to state-of-the-art classical and recent learned SfM pipelines.

\begin{figure*}[t]
    \centering
    \includegraphics[width=\linewidth]{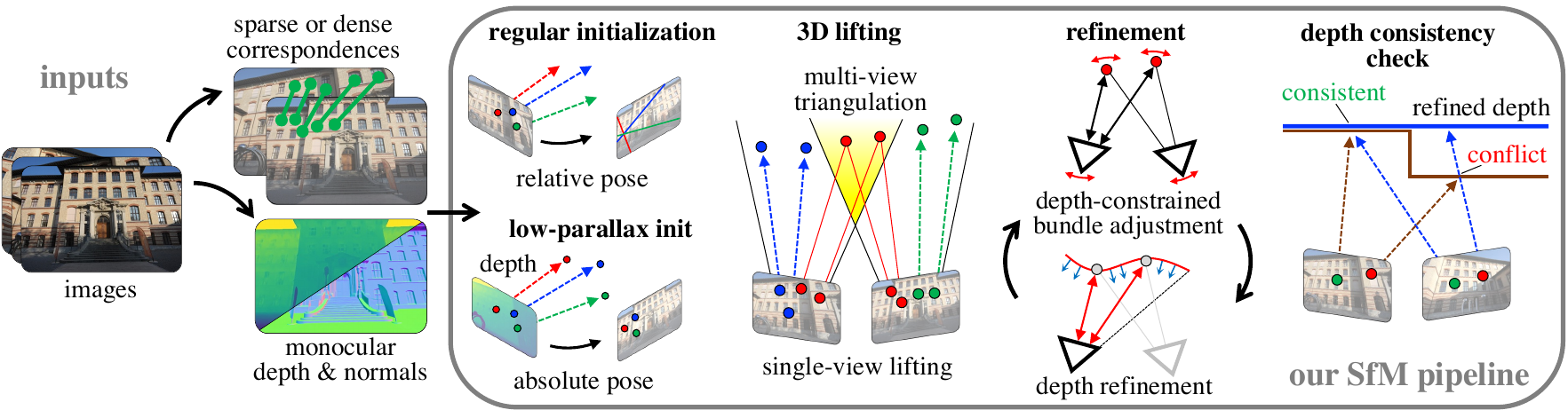} 
    \caption{\textbf{Overview of our approach.}
    Given image correspondences, depth, and surface normals, we first initialize the reconstruction by estimating a relative pose or, if the parallax is low, an absolute pose from points lifted to 3D by depth. 
    While SfM can generally estimate 3D only for points observed in multiple views, we leverage single-view observations with depth.
    This helps registering images with lower visual overlap.
    Camera poses, 3D points, and depth maps are refined by alternating between bundle adjustment and normal integration with depth constraints.
    Finally, we reject incorrect registrations, \eg, due to symmetries, by checking that the depth is consistent across views.
    }%
    \label{fig:architecture}%
\end{figure*}

\section{Related work}

\paragraph{Traditional SfM:}
In the early days, the field of computer vision largely focused on SfM from ordered video sequences~\cite{beardsley1997sequential,szeliski1994recovering,pollefeys2004visual} while later works shifted to unordered image collections~\cite{schaffalitzky2002multi,snavely2006photo,agarwal2011building,wu2011visualsfm,frahm2010building,heinly2015reconstructing,schoenberger2016sfm}.
The literature has traditionally categorized methods into the incremental and global paradigms.
Over the years, several software packages have become available~\cite{snavely2006photo,wu2011visualsfm,moulon2016openmvg,schoenberger2016sfm,pan2024glomap} with COLMAP as the, arguably, most widely adopted SfM pipeline.

While achieving reliable results for a wide range of inputs, each pipeline exhibits its own unique weaknesses and failure modes.
Common to all approaches, however, is the fundamental requirement for three-view overlap and tracks, which we address specifically in this work.
Prior works~\cite{josephson2007image,camposeco2018hybrid} have identified this issue as well and proposed methods for camera pose estimation from both hybrid 2D-3D and 2D-2D correspondences.
Sinha~\etal~\cite{sinha2004camera} presented an approach for pure two-view SfM from silhouettes.
However, it relied on the extraction of silhouettes in outside-in capture scenarios, while we intend to solve more general scenarios.
Furthermore, it requires known epipolar geometry to at least two previously registered views and thus still needs three-view overlap.
Zheng and Wu~\cite{zheng2015structure} also tackled the issue and proposed an approach for structure-less resectioning from 2D-2D correspondences.
While this does away with needing three-view tracks, equivalent to Sinha~\etal~\cite{sinha2004camera}, it still requires three-view overlap (\ie, three images are all pairwise matchable but no two-view matches form three-view tracks) to constrain the scale.
It also suffers from common degenerate viewpoint configurations, such as sideward motion.
In contrast, we require neither three-view tracks nor three-view overlap and also do not suffer from the same degeneracies due to integration of strong monocular priors.

\paragraph{Learning for SfM:}
Driven by the overwhelming success of machine learning, many works have been devised to integrate data-driven methods into individual components of the traditional pipeline with a focus on addressing the challenges in the feature representation~\cite{Schonberger2017Comparative,mishchuk2017working,superpoint,dusmanu2019d2} and matching~\cite{sarlin2020superglue,lindenberger2023lightglue,sun2021loftr,dust3r_cvpr24,dust3r_cvpr24,dusmanu2020} stages.
Relatively fewer works tackled other components like RANSAC~\cite{brachmann2021dsacstar,wei2023generalized}, bundle adjustment~\cite{lindenberger2021pixsfm,wei2020deepsfm}, or camera calibration~\cite{veicht2024geocalib}.
Notably, the recent works of DuSt3R~\cite{dust3r_cvpr24} and MASt3R~\cite{leroy2024grounding} demonstrate impressive two-view matching and reconstruction results.
Based on this feat, the MASt3R-SfM~\cite{duisterhof2024mast3r} pipeline performs multi-view reconstruction using a paradigm akin to traditional global SfM.
Furthermore, several methods take advantage of an end-to-end learning objective to jointly train multiple SfM components~\cite{brachmann2024acezero,wang2023vggsfm,smith2024flowmap}.
Despite achieving state-of-the-art results in specific scenarios, these pipelines still do not serve as universal replacement for traditional SfM in more general, unstructured, or large-scale settings~\cite{brachmann2024acezero,wang2023vggsfm,smith2024flowmap}.
In contrast, we integrate depth priors into traditional incremental SfM to solve its failure modes while retaining the generality and scalability of the classical approaches.

\paragraph{Monocular depth priors:}
Since the early works on monocular depth estimation~\cite{saxena2005learning,hoiem2005geometric,eigen2014depth}, astounding progress has been made.
The latest models~\cite{yin2023metric3d,ke2023repurposing,depth_anything_v2,depth_pro,bae2024dsine} are able to estimate depths and normals with zero-shot generalization on in-the-wild images.
Typically, the resulting depth maps are visually pleasing with a sharp delineation of occlusion boundaries and good relative depth accuracy.
When it comes to directly using the depth estimates for metric, multi-view 3D reconstruction, the absolute accuracy is typically insufficient and inconsistent across posed overlapping views.

In our work, we build on the tremendous progress to improve upon the limitations of SfM.
Our method leverages monocular depth to reconstruct previously unsolved two-view overlap scenarios as well as to improve the overall robustness of the system.
Through a carefully designed optimization strategy and principled uncertainty propagation, we can robustly deal with noisy monocular depth input and jointly refine them with the multi-view triangulated structure.
We use monocular surface normals, which are easier to predict than depth and often more accurate, by integrating them into depth following Cao~\etal~\cite{cao2022bilateral}.
COMO~\cite{dexheimer2024como} also leverages normal integration and monocular depth for real-time monocular mapping and odometry with a tightly coupled system, similar to ours.
However, their approach is designed for real-time operation and is thus limited to GPU and to a few images.
Furthermore, it is not amenable to the more general problem of SfM from unstructured images.

Closer to our approach, StudioSfM~\cite{studiosfm} leverages monocular depth in SfM but only for the 
scenario of low-parallax videos by modifying COLMAP's initialization and bundle adjustment stages.
Ours is more general and tackles multiple of the limitations of incremental SfM, including low-parallax, to handle unstructured image collections.

\begin{figure*}[t]
    \centering
    \setlength{\pheight}{0mm}
    \setlength{\pwidth}{0.005\linewidth}
    \setlength{\iwidth}{0.35\linewidth}
    \setlength{\lwidth}{\dimexpr(0.999\linewidth - 2\pwidth - \iwidth)/2 \relax}

    \includegraphics[width=\iwidth]{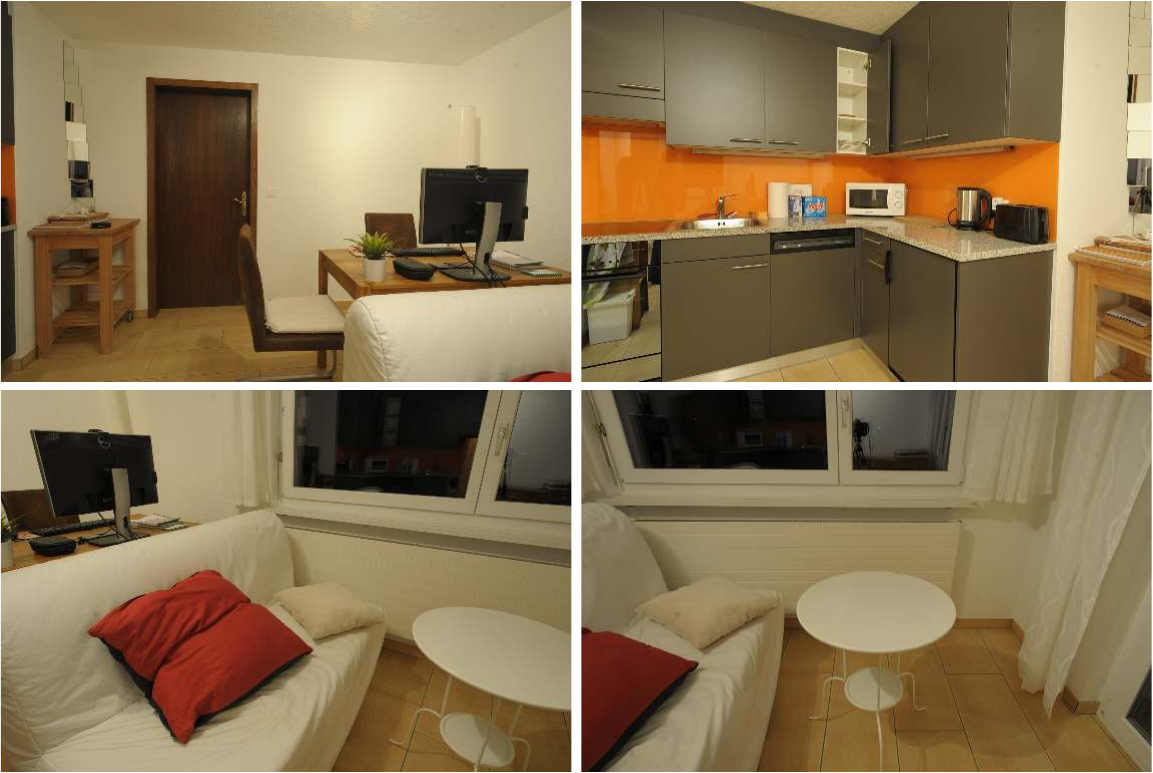}%
    \hspace{\pwidth}%
    \includegraphics[width=\lwidth]{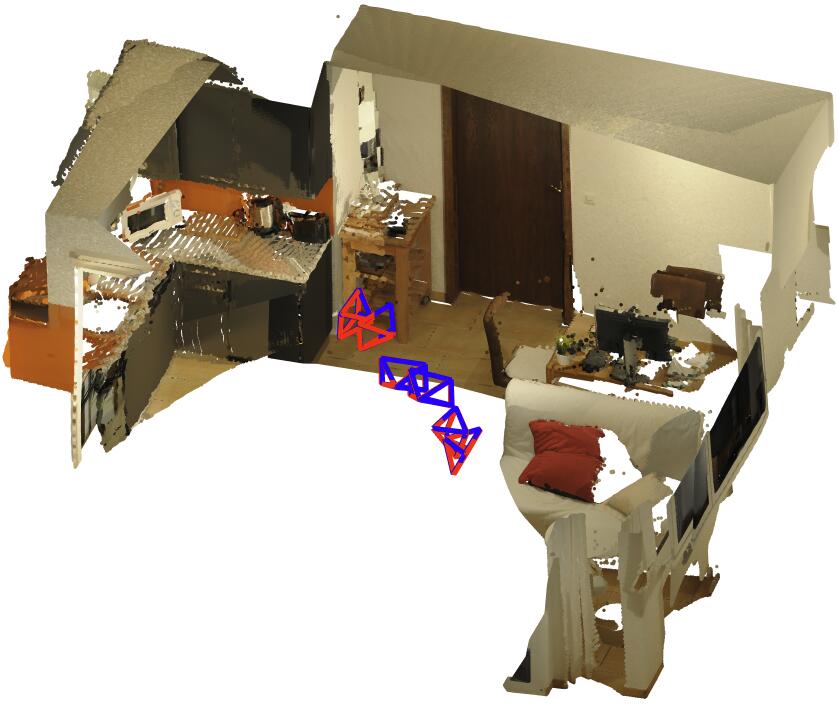}%
    \hspace{\pwidth}%
    \includegraphics[width=\lwidth]{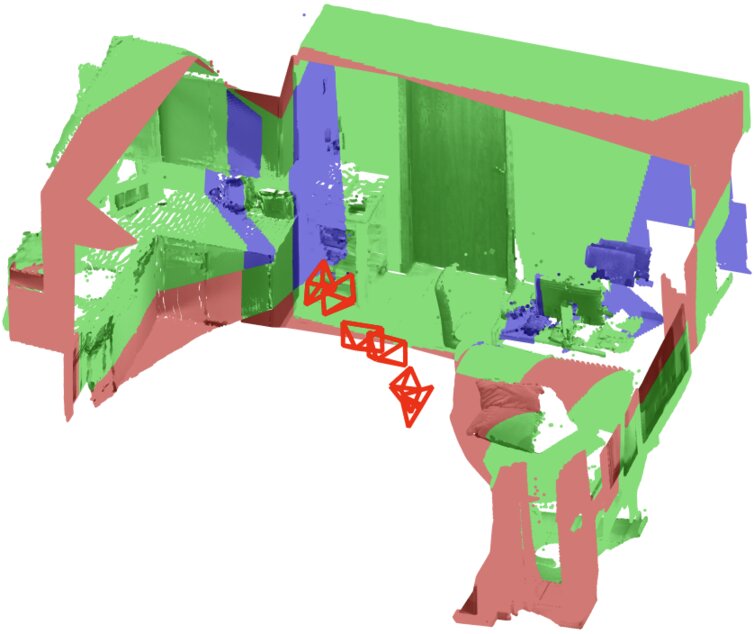}%
    \vspace{\pheight}
    
    \includegraphics[width=\iwidth]{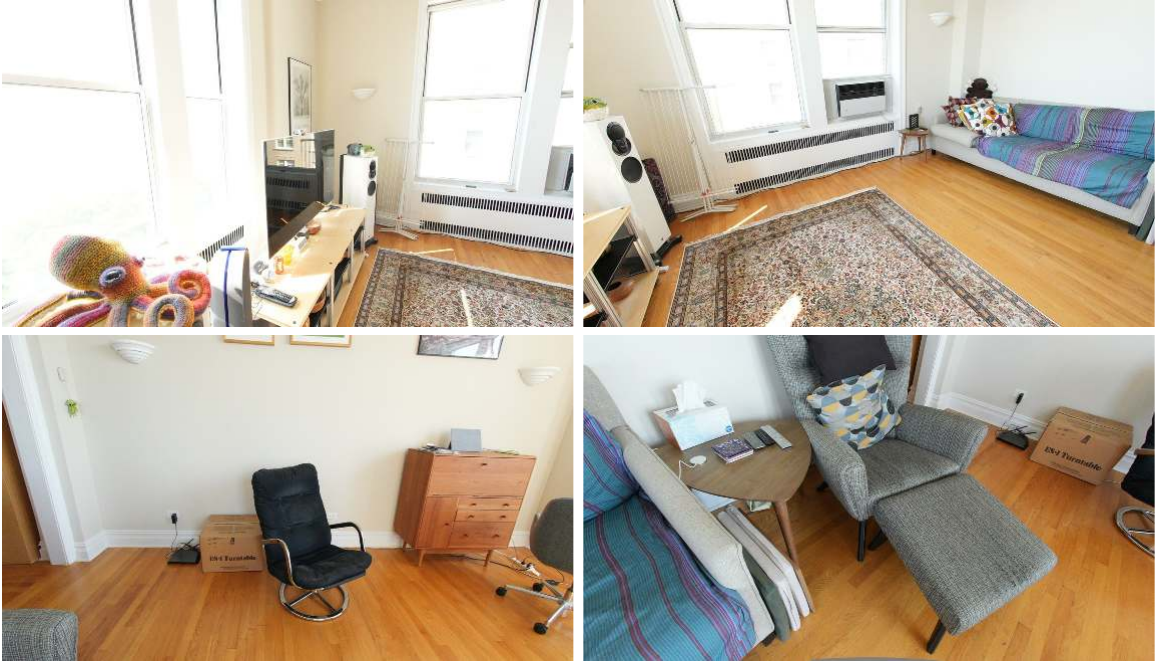}%
    \hspace{\pwidth}%
    \includegraphics[width=\lwidth]{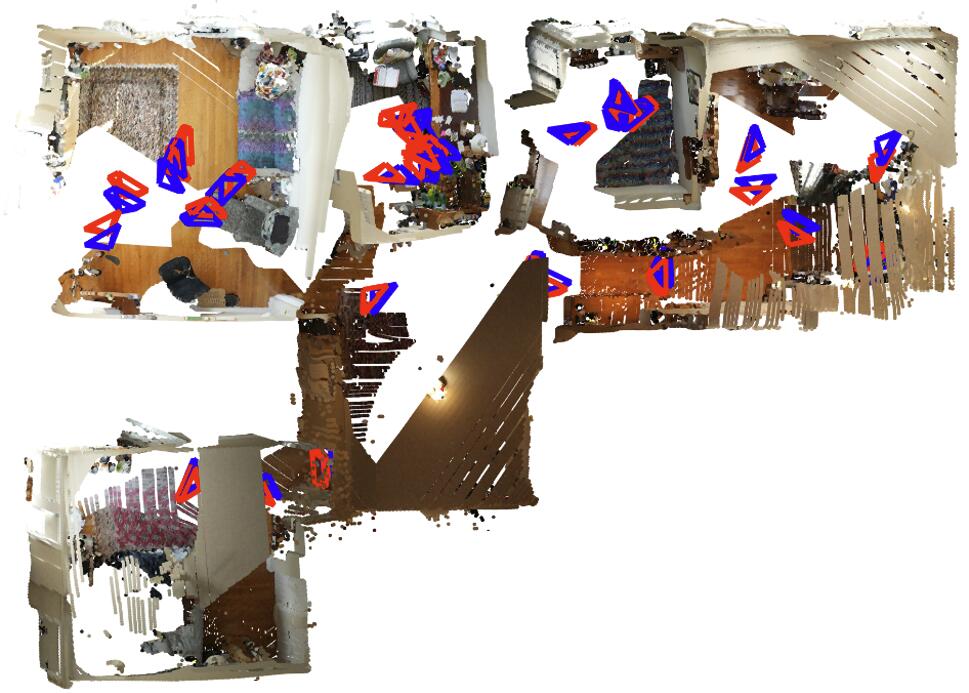}%
    \hspace{\pwidth}%
    \includegraphics[width=\lwidth]{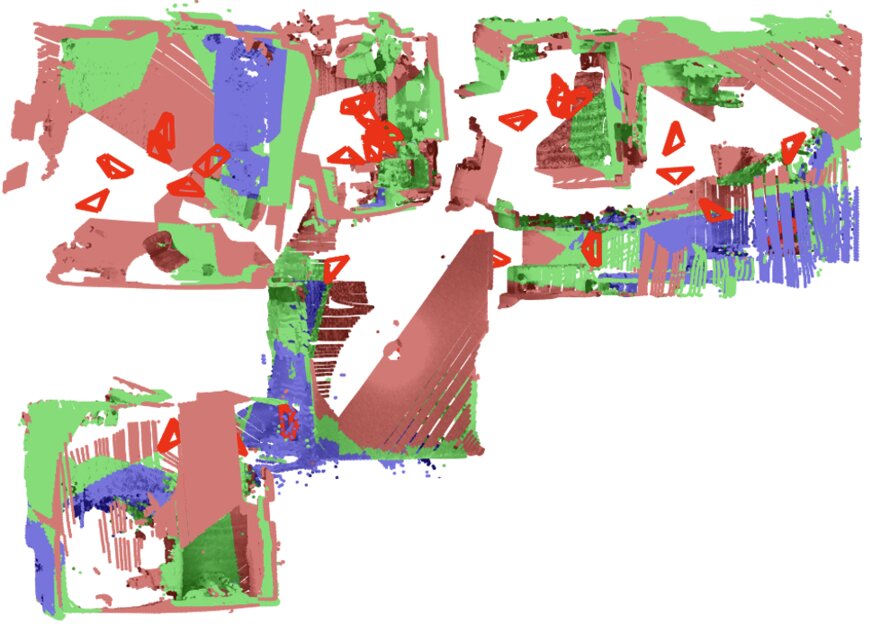}%
    \vspace{\pheight}
    
    \includegraphics[width=\iwidth]{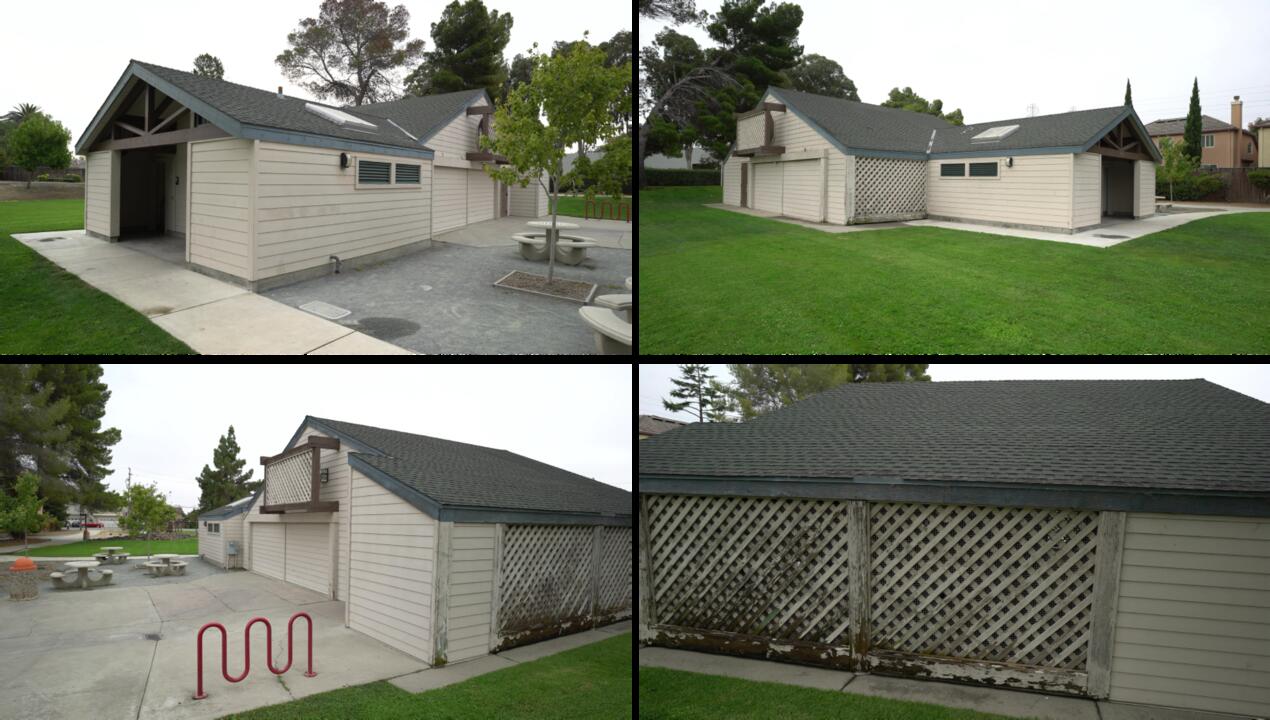}%
    \hspace{\pwidth}%
    \includegraphics[width=\lwidth]{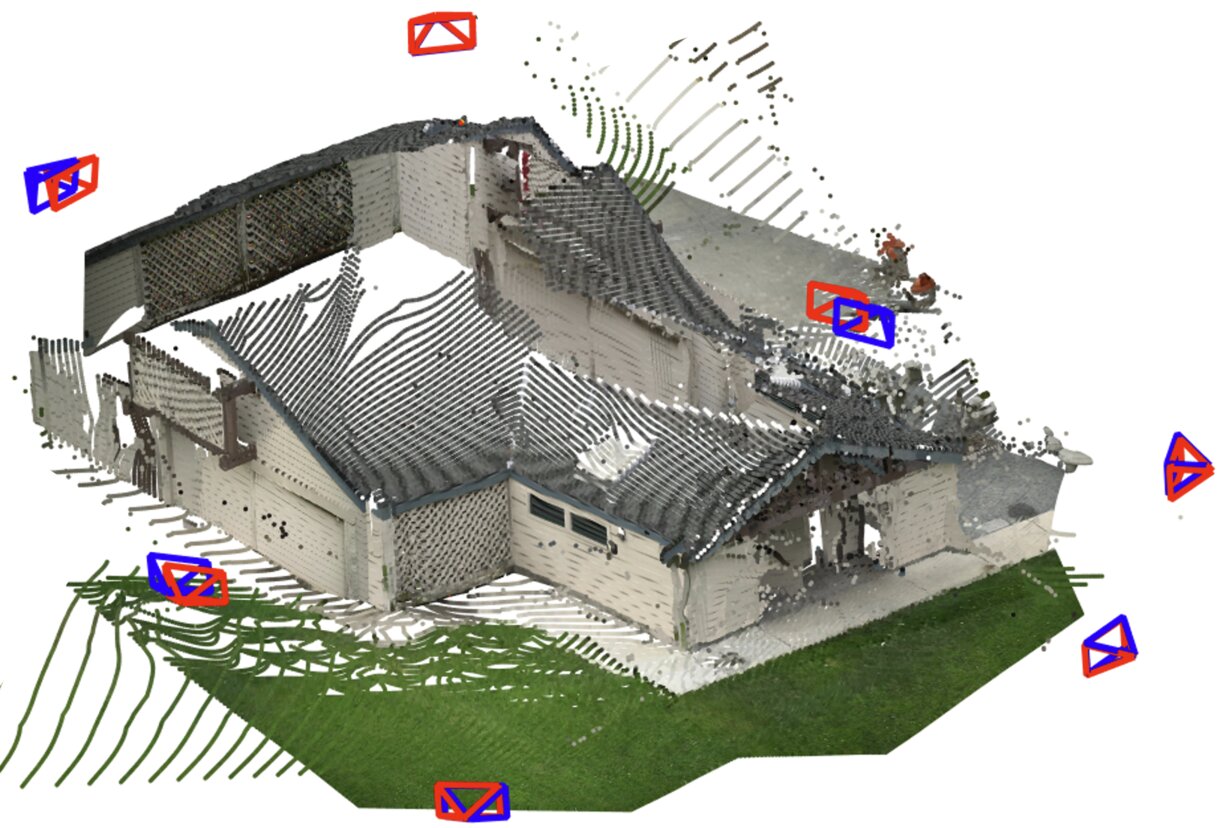}%
    \hspace{\pwidth}%
    \includegraphics[width=\lwidth]{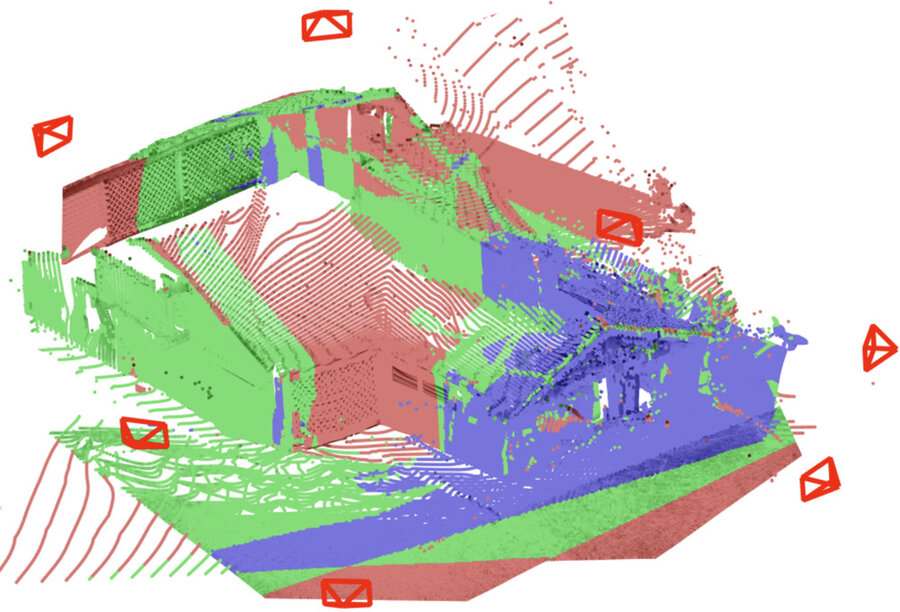}%
    \vspace{\pheight}

    \caption{\textbf{Qualitative results for low overlap scenes.}
    Left: Input images with low overlap.
    Center: \red{Estimated (red)} and \blue{ground-truth (blue)} camera poses with the monocular depth refined by our system.
    Right: Lifted refined depth of which points are colored differently whether they are visible in \red{a single image (red)}, {\color[HTML]{33cc33}two (green)}, or \blue{at least three images}.
    }%
    \label{fig:qualitative}%
\end{figure*}

\section{Method}

We first formulate the problem and provide an overview of our system -- see also \cref{fig:architecture}. 

\paragraph{Inputs:}
Our system takes as input a set of $i = 1...n$ unordered images $\mathcal{I} {=} \{ I_i \in \real^{H_i \times W_i} \}$ and their intrinsic pinhole camera parameters $\mathcal{K} {=} \{ K_i \in \real^{3 \times 3} \}$.
For each of them, we estimate monocular depth maps $\mathcal{D} {=} \{ D_i \in \real_0^{+{H_i \times W_i}} \}$ and normal maps $\mathcal{N} {=} \{ N_i \in \mathcal{S}^{2^{H_i \times W_i}} \}$ with their respective confidence maps $\Sigma_{D_i}, \Sigma_{N_i}$.

Following standard practice, we extract sparse local image features $\mathcal{F} {=} \{ \mathcal{F}_i \}$ for each image, where features $\mathcal{F}_i$ in image $i$ are defined by their pixel keypoint positions $\{ x_j \in [0, W_i] {\times} [0, H_i]\ |\ j = 1...f_i \}$ with assumed Gaussian noise $\Sigma_{x_j}$ {and descriptors $\{ \bm{d}_j \in \real^d \}$}.
Feature matching and geometric verification produce correspondences $\mathcal{M} {=} \{ \{\mathcal{M}_{a,b}\} \subset [1...f_a] {\times} [1...f_b] \}$ with associated scores 
$\{ q_{a,b} \in \real^{|\mathcal{M}_{a,b}|} \}$, between sparse features for each image pair, which later form feature tracks across multiple images.
We optionally compute dense pixel-wise matches, 
sub-sample them using either sparse keypoints or non-maximum suppression, and apply geometric verification, resulting in two-view correspondences
$\mathcal{M}^* {=} \{ \{\mathcal{M}^*_{a,b}\} \subset [1...H_a W_a] {\times} [1...H_b W_b] \}$ with scores $ \{ q^*_{a,b} \in \real^{|\mathcal{M}^*_{a,b}|} \}$.

\paragraph{Outputs:}
Given the intrinsics $\mathcal{K}$, depth maps $\mathcal{D}$, normal maps $\mathcal{N}$, sparse features $\mathcal{F}$ as well as correspondences $\mathcal{M}$ and $\mathcal{M}^*$, our incremental reconstruction algorithm estimates the camera poses $\mathcal{P} {=} \{ P_i \in \mathrm{SE}(3)\ |\ i \in \registered\}$ for 
a subset $\registered$ of images that could be confidently registered.
It also estimates $k = 1...m$ scene points $\mathcal{X} {=} \{ X_k \in \real^3 \}$ and refined, globally consistent depth maps $\mathcal{D}^*$.

\paragraph{System overview:}
Our system builds upon the COLMAP incremental SfM framework~\cite{schoenberger2016sfm}.
COLMAP only takes prior calibrations $\mathcal{K}$, sparse features $\mathcal{F}$, and correspondences $\mathcal{M}$ to estimate the output camera poses $\mathcal{P}$ and scene points $\mathcal{X}$.
To leverage the monocular priors $\mathcal{D}$, $\mathcal{N}$ and the dense two-view correspondences $\mathcal{M}^*$, we make significant changes to many of the underlying algorithms as well as the control logic in the pipeline.
The correspondence graph is built from either sparse or dense correspondences.
Then, similar to COLMAP, we start with an initial image pair and incrementally register more images, interleaved with local and global refinements.

\subsection{Two-View Initialization}
\label{sec:two-view-initialization}

\paragraph{Initial pose:}
Following COLMAP, 
we rank image pairs by number of inlier correspondences and
select the first pair $(I_a, I_b)$ 
yielding a stable relative pose $T_{ba}$ $\in \mathrm{SE}(3)$, \ie, sufficient inliers and parallax. If no such pair exists, we leverage the monocular depth prior to constrain the cameras.
We lift feature points from $I_a$ using its depth $D_a$ and calibration $K_a$, and form 2D-3D matches with corresponding points in $I_b$ to estimate the absolute pose $T_{ba}$ via PnP~\cite{chum2003locally,Haralick94IJCV}.
We initialize the reconstruction using the estimated pose 
\ie, $P_b =$ $T_{ba}$ and $P_a$ as the identity.

\paragraph{Initial 3D points:}
We initialize scene points $\mathcal{X}$ by lifting low-parallax inliers and triangulating the rest.
Next, we scale each depth map $D_i$ to be consistent with these 3D points by computing a scaling factor 
\begin{equation}
D_i^* = D_i\cdot
     \underset{j,k}{\mathrm{median}}
    \left(\left\{{\hat{D}_i(X_k)}/{D_i(x_j)}\right\}\right)
    \enspace,
    \label{eq:depth-scaling}
\end{equation}
where $\hat{D}_i(X_k) = (P_i \cdot X_k)_3$ denotes the depth of point $X_k$ in camera frame $i$ and $D_i(x_j)$ interpolates the depth map at coordinate $x_j$.
We then refine the depth maps and the 3D points following the optimization described later in~\cref{sec:local-global-refinement}.

Next, we check whether the resulting depth maps are consistent, as described in \cref{sec:depth-consistency-check}, to reject the initial image pair if it is incorrectly posed, \eg, due to symmetry.
If they are consistent, we accept the pair and consider the images as registered, \ie, $\registered = \set{a, b}$, otherwise, we search for another pair.
Finally, we augment the scene points $\mathcal{X}$ with image points not yet associated with a scene point by lifting them using their respective scaled and refined monocular depth.

\subsection{Next View Registration}
\label{sec:next-view-registration}

\paragraph{View selection:}
To register a next view, we consider images that have not been registered before, \ie, $\{ I_c\ \vert\ c\not\in \registered\}$. %
We rank these candidates by their maximum 
two-view 
correspondence scores to registered images, \ie, $\arg\max_{i \in \registered} \left( \sum q_{c,i} + \sum q^*_{c,i} \right)$.
In ranked order, we attempt to register the candidates using 2D-3D correspondences and robust absolute pose estimation.
Here, the 2D-3D correspondences contain both regular multi-view triangulated 3D points but also single-view 3D points previously lifted using monocular depth.
This enables us to register a next view without three-view overlap, whereas other SfM approaches can only consider points that have been triangulated from at least two registered views (see \cref{fig:qualitative} and Appendix {\color{red}H.3}).

\paragraph{Registration:}
If we find enough inliers, we extend the reconstruction with the estimate of the camera pose $P_c$.
After scaling its depth map following~\cref{eq:depth-scaling}, we extend the structure by either continuing the tracks of existing points or initialize new ones from lifted single-view points.
Next, we refine the reconstruction (\cref{sec:local-global-refinement}) and check the depth consistency (\cref{sec:depth-consistency-check}).
If successful, we label the image as registered, \ie, $\registered \cup \set{c}\rightarrow\registered$.
Otherwise, we discard the new image with its associated structure and we try a different image.

\subsection{Local and Global Refinement}
\label{sec:local-global-refinement}

After two-view initialization and registration of new images, we jointly refine the camera poses and the scene structure.
Following COLMAP's scheduling for local and global bundle adjustment, we perform either a global refinement over all registered images and scene points or over a local window around the last registered image.
This strategy results in an amortized linear runtime for incremental SfM~\cite{wu2013towards}.

\paragraph{Optimization problem:}
Different from standard bundle adjustment, we condition the estimated scene structure and camera poses on the monocular depth and normal priors.
To robustly deal with noisy priors, which are generally predicted by imperfect neural networks, we estimate a set of multi-view consistent depth maps $\mathcal{D}^*$.
We model this objective by optimizing the following overall cost function
\begin{equation}
   \arg \min_{\mathcal{P}, \mathcal{X}, \mathcal{D}^*} C_{\mathrm{BA}} + C_{\mathrm{reg}} + C_{\mathrm{int}} \enspace.
\end{equation}
The first term defines the standard bundle adjustment cost function for each observation of multi-view 3D points as
\begin{equation}
   C_{\mathrm{BA}} = \sum_{i\in\registered} \sum_{j,k} \rho_{\mathrm{BA}} 
   \left( \norm{ \pi\left(K_i, P_i, X_k\right) - x_{j} }^2_{\Sigma_{x_j}}  \right) \ ,
\end{equation}
where $\norm{\cdot}_\Sigma$ is the Mahalanobis distance, $\rho_{\mathrm{BA}}$ is a truncated Smooth-$L1$ loss, and $\pi(K, P, X) \in \real^2$ projects  scene 
points into the image plane in pixel units.
The second term penalizes deviations between scene points and the refined depths:
\begin{equation}
   C_{\mathrm{reg}} = \sum_{i\in\registered} \sum_{j,k} \rho_{\mathrm{reg}} 
   \left( \norm{ \hat{D}_i(X_k) - D_i^*\left( x_j \right) }^2  \right) \ ,
\end{equation}
which is robustified with the Cauchy loss $\rho_{\mathrm{reg}}$. Appendix {\color{red}H.2} shows supporting examples.
The last term defines a depth integration cost that conditions the refined depth maps on the monocular depth and normal priors
\begin{align}
\label{eq:depth integration cost}
   C_{\mathrm{int}} = &\sum_{i\in\registered} \sum_{u,v}\left[ 
   \rho_{\mathrm{prior}} \left( \norm{D_i^*(u, v) - D_i(u, v)}^2_{\Sigma_{D_i}(u, v)} \right)\right. \nonumber \\
   & + \left.\rho_{\mathrm{int}} \left( \norm{N_i(u, v) - \Delta D^*_i(u, v)}^2_{\Sigma_{N_i}(u, v)} \right)\right] \ ,
\end{align}
where $(u, v) \in [0, W_i] \times [0, H_i]$ are pixel coordinates, $\Delta D_i^*(u, v) \in \mathcal{S}^2$ defines the surface tangent plane normal using first-order differentiation of depth map $D_i^*$ at $(u, v)$, and $\rho_\mathrm{prior,int}$ are truncated $L_2$ robust losses.
In the depth integration, the first term conditions the refined on the prior depth, while the second term uses bilateral normal integration~\cite{cao2022bilateral} with uncertainty weighting, for which we provide the exact formulation in Appendix {\color{red} B}.

\paragraph{Efficient solving:}
The joint objective over $C_{\mathrm{BA}} + C_{\mathrm{reg}} + C_{\mathrm{int}}$ leads to a sparsity structure in the Hessian, illustrated in Appendix {\color{red}G}, not amenable to the Schur complement trick.
To retain a tractable optimization, we therefore use an alternating block coordinate descent strategy, where we first minimize $C_{\mathrm{reg}} + C_{\mathrm{int}}$, for each image independently, using fixed scene points $\mathcal{X}$ and their propagated depth covariances $\Sigma_{\hat{D}_i(X_k)}$.
Then, using fixed refined depth maps $\mathcal{D}^*$ and their propagated covariances $\Sigma_{D_i^*(x_j)}$, we jointly minimize $C_{\mathrm{BA}} + C_{\mathrm{reg}}$ over all views $\mathcal{P}$ and scene points $\mathcal{X}$.
We alternate between these two steps for several rounds and interleave careful filtering of scene points using COLMAP's reconstruction filtering strategy.

\begin{figure}[t]
    \centering
    \setlength{\pheight}{1mm}
    \setlength{\bwidth}{2.5cm}
    \setlength{\pwidth}{0.005\linewidth}
    \setlength{\iwidth}{\dimexpr(0.999\linewidth - 2\pwidth)/2 \relax}

    \includegraphics[width=\linewidth]{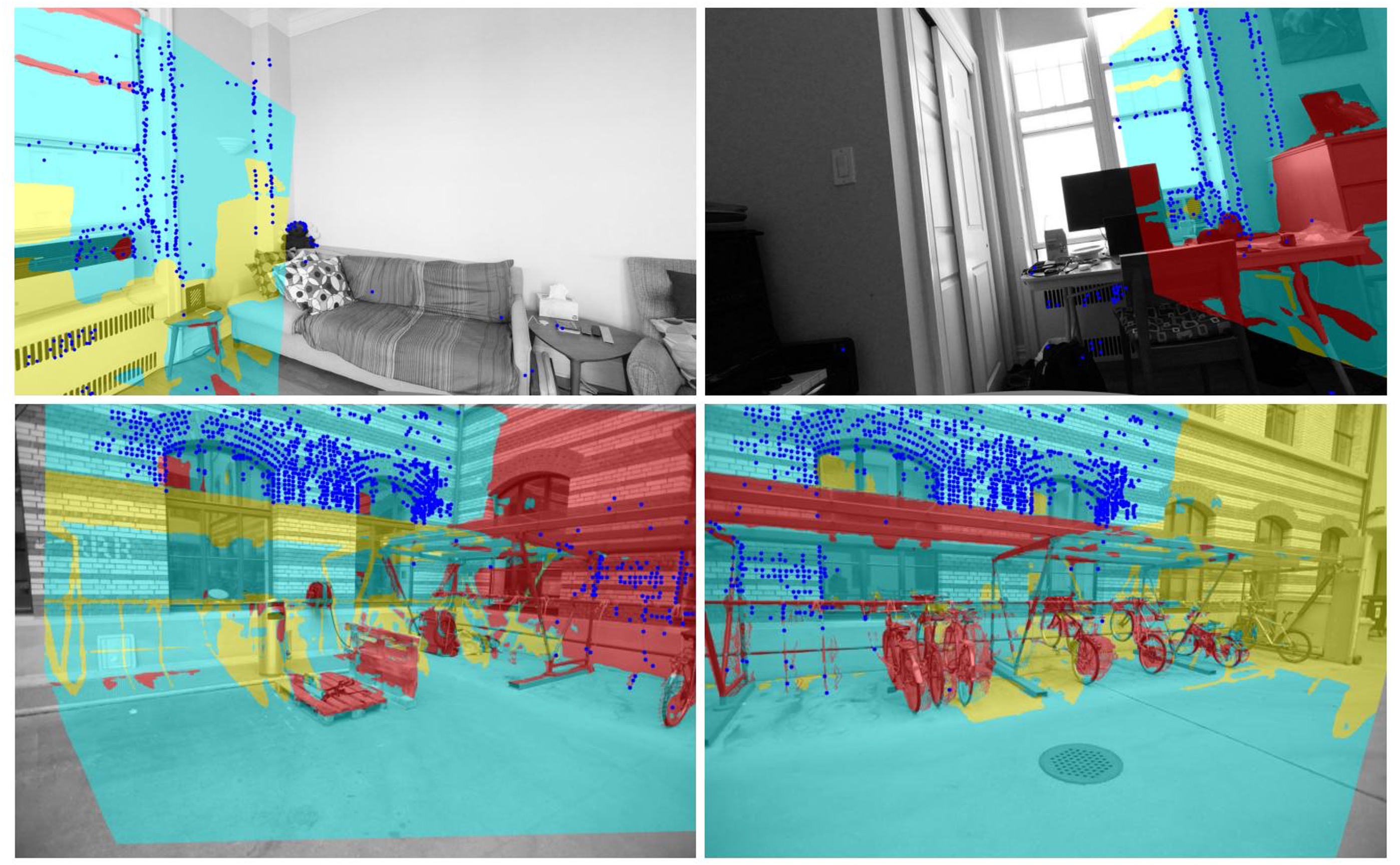}%

    \caption{\textbf{Depth consistency check.} 
    These two image pairs are incorrectly matched because of symmetries ({\color{blue} blue points}).
    Our approach successfully rejects them as a large ratio of pixels have an inconsistent depth ({\color{red}red}), while ignoring occlusion ({\color{Goldenrod}yellow}) and areas with consistent depth ({\color{cyan}cyan}).
    }%
    \label{fig:depthcheck}%
\end{figure}

\begin{table*}[t]
\centering
\scriptsize
\setlength\tabcolsep{2pt}%
\newcommand{\colordot}[1]{{\large $\color[HTML]{#1}\bullet$}}%
\renewcommand{\b}[1]{\textbf{#1}}
\begin{minipage}{0.39\linewidth}
\includegraphics[width=\textwidth]{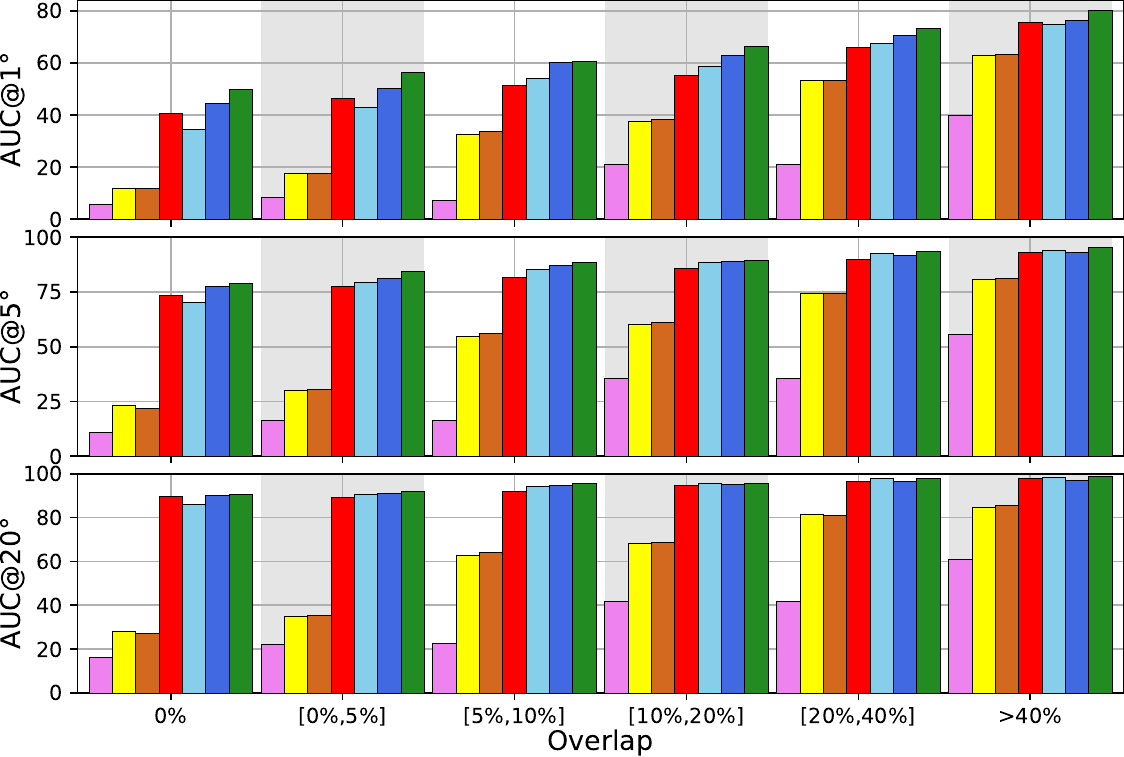}%
\end{minipage}%
\hspace{2mm}%
\begin{tabular}{llccccc}
\toprule
\multirow{2}{*}[-0.4em]{matching} & \multirow{2}{*}[-0.4em]{approach}
& \multicolumn{4}{c}{three-view visual overlap} & \multirow{2}{*}[-0.4em]{\makecell{all images\\max overlap}}\\
\cmidrule(lr){3-6}
& & minimal, 0\% & $<$5\% & $<$10\% & $<$30\%\\
\midrule
{\large $\color[HTML]{ee82ee}\bullet$} SIFT & COLMAP
& \01.1/\02.3/\03.0 & \01.1/\02.3/\03.1 & \02.3/\03.8/\04.4 & \06.5/\08.8/\09.4 & 63.1/77.2/80.7 \\

\colordot{ffff00} SP+LG & COLMAP
& \07.9/12.7/14.6 & \06.9/12.5/15.4 & 12.0/19.8/23.2 & 44.9/60.3/65.0 & 67.2/80.6/83.9 \\
\colordot{d2691e} SP+LG & SLR
& \08.2/13.1/15.2 & \06.7/12.6/15.7 & 11.2/18.3/21.3 & 46.1/62.2/67.1 & 67.7/81.3/84.7\\

\hspace*{2mm} SP+LG & GLOMAP
& \08.4/15.8/22.5 & \05.5/15.3/25.7 & 12.1/25.3/35.7 & 50.1/66.7/71.8 & 67.5/78.5/82.3\\
\colordot{ff0000} SP+LG & \bf{ours}
& 27.3/55.9/71.8  & 26.4/55.2/71.3  & 30.0/56.4/70.1 & 57.0/79.1/86.0  & \textbf{74.3}/\textbf{88.3}/92.0 \\

\midrule
\multicolumn{2}{c}{VGG-SfM} %
& \00.9/\03.2/\05.1 & \00.6/\02.7/\05.0 & \01.8/\05.5/\08.6  & \07.8/19.6/25.7 & 27.7/52.5/61.8\\

\hspace*{2mm} RoMa & COLMAP
& \07.3/11.9/14.0 & \05.3/9.6/12.4 & 11.5/18.9/22.6 & 41.4/56.4/61.8 & 63.6/76.4/79.7

\\
\hspace*{2mm} LoFTR & DF-SfM
& \03.2/\07.9/10.7  & \02.2/\06.8/10.1 & \04.0/11.6/15.9  & 25.6/51.1/60.2 & 54.9/80.0/85.9\\

\colordot{87ceeb} MASt3R & M-SfM
& 20.1/39.7/52.2 & 17.2/37.5/49.2 & 19.8/37.3/48.1 & 31.4/50.4/59.2 & 50.5/67.9/74.1 \\

\colordot{4169e1} MASt3R & \bf{ours} 
& \textbf{34.9}/\textbf{67.2}/\textbf{81.7} & \textbf{34.0}/\textbf{67.2}/\textbf{81.2}  & 37.7/\textbf{67.8}/\textbf{80.6}  & 55.5/79.3/\textbf{86.6}  & 70.3/88.2/\textbf{93.6} \\

\colordot{228b22} RoMa & \bf{ours}
& 33.4/60.6/74.4 & 32.8/60.7/74.2 & 39.7/65.8/77.9 & \textbf{60.3}/\textbf{79.6}/85.6 & 71.6/87.0/91.3 \\
\bottomrule

\end{tabular}
\caption{\textbf{SfM with low overlap on the ETH3D dataset.}
Left: We construct a minimal dataset of triplets with reduced three-view  overlap, as in~\cref{fig:teaser}.
Right: We consider larger sets of images with increasing overlap.
We report the AUC of the pose error up to 1/5/20\degree.
Our approach yields the most accurate poses given either sparse or dense correspondences.
\label{tbl:results:eth3d}%
}
\end{table*}

\begin{table*}[t]
\centering
\scriptsize
\setlength\tabcolsep{5.5pt}%
\renewcommand{\b}[1]{\textbf{#1}}
\begin{tabular}{llcccccccc}
\toprule
&& \multicolumn{4}{c}{SMERF dataset} & \multicolumn{4}{c}{Tanks \& Temples dataset}\\
\cmidrule(lr){3-6} \cmidrule(lr){7-10}
matching & overlap $\rightarrow$ & minimal & low & medium & high & minimal & low & medium & high\\
\midrule
SIFT & COLMAP
&\0.1/\00.2/\00.2 & \00.0/\00.0/\00.0 & \00.0/\00.0/\00.0 & \06.1/\07.6/\08.1 %
& \00.6/\03.2/\05.3 & \00.2/\02.4/\04.7 & \00.6/\04.3/\06.8 & \04.1/21.1/30.0 %
\\
SP+LG & COLMAP
& 2.2/4.2/4.9  & \03.4/7.8/9.9  & 17.7/30.7/36.0 & 42.9/55.4/59.5 %
& \02.0/11.9/18.2 & \02.0/11.9/18.2 & \02.0/11.9/18.2 & 15.7/52.9/66.8 %
\\
SP+LG & SLR
& 2.3/4.2/\05.0 & \03.7/\08.6/10.9\ & 16.5/29.2/34.5&  43.0/56.3/60.7 %
& \02.0/12.7/19.5 & \03.8/19.4/29.5 & \08.6/37.3/50.5  & 15.9/53.7/68.0 %

\\
SP+LG & GLOMAP
& 3.5/8.0/13.3  & 3.4/9.6/15.9 & 20.9/35.3/42.8 & 48.1/60.9/65.4 %
& \01.3/9.1/17.8 & \02.7/15.7/25.0 & \09.0/30.9/39.9 & 16.1/45.7/55.9 %

\\
SP+LG & \bf{ours}
&  9.2/41.0/69.8  & 5.4/29.1/53.0 & 14.0/47.6/72.9 & 47.3/79.3/90.6 %
& \05.0/29.2/54.9 & \05.0/34.1/58.1 & 10.8/51.3/72.8 & 18.9/62.8/80.9 %

\\
\midrule
\multicolumn{2}{c}{VGG-SfM}
& \00.0/\00.1/\00.3 & \00.0/\00.0/\00.3 & OOM & OOM %
& \01.8/11.0/20.0 & \08.2/39.2/55.4 &  13.7/49.1/64.9 & 19.3/56.8/70.7 %
\\
RoMa & COLMAP 
& 2.5/4.3/5.2 & 6.9/13.5/16.3 & 23.0/36.2/40.7 & 40.5/50.2/53.1 %
& \03.2/13.6/19.6 & \07.4/28.5/38.7 & \012.8/43.0/54.1 & 17.4/46.4/56.9 %
\\
LoFTR & DF-SfM
& 0.5/\01.2/\01.6 & \00.7/\01.5/\02.1 & \03.8/\08.6/11.1 & 20.6/33.6/39.6 %
& \01.4/12.0/20.9 & \01.4/12.0/20.9 & \04.9/28.5/42.2 & 13.4/51.7/66.7 %
\\
MASt3R & M-SfM
& 3.9/10.4/18.0 & \04.3/11.7/23.0 & \05.9/15.8/28.1 & 10.4/22.9/39.9 %
& \textbf{18.0}/53.5/68.6 & \textbf{26.6}/61.8/72.0 & \textbf{27.0}/62.5/73.0 & \textbf{28.5}/64.4/75.6 %
\\

MASt3R & \bf{ours} 
& \textbf{17.2}/\textbf{54.6}/\textbf{77.1} & \textbf{26.6}/\textbf{63.2}/\textbf{84.1}  & \textbf{40.4}/\textbf{72.8}/\textbf{87.5} & \textbf{57.1}/\textbf{84.6}/\textbf{94.1}  %
& 14.8/\textbf{56.8}/\textbf{79.6} & 21.7/\textbf{65.7}/\textbf{83.4} & 22.8/\textbf{68.4}/\textbf{85.5} & 25.9/\textbf{71.1}/\textbf{86.7} %
\\
RoMa & \bf{ours}
& 10.6/41.0/61.8 & 10.1/37.2/59.2  & 21.1/48.4/66.7 & 41.4/69.3/79.4 %
& 7.2/41.5/65.3 & 10.3/44.0/63.2  & 14.2/56.2/75.9 & 22.1/67.4/86.1 %
\\
\bottomrule
\end{tabular}%
\caption{\textbf{SfM with low overlap on the SMERF and T\&T datasets.}
We report the AUC of the pose error up to 1/5/20\degree\ for approaches based on sparse (top) and dense matching (bottom).
Ours is the only approach capable of reconstructing low-overlap scenes in the SMERF dataset.
\label{tbl:results:sfm}%
}
\end{table*}

\subsection{Depth Consistency Check}
\label{sec:depth-consistency-check}

Filtering the reconstruction based on sparse image observations alone, as performed by COLMAP and others, is important but limits the ability for identifying conflicting observations through occlusion and free-space reasoning~\cite{merrell2007real,heinly2014correcting,taira2019right}.
As such, traditional sparse SfM often fails catastrophically when a single image is incorrectly registered due to symmetry issues, estimation failures, or other wrong decisions.

Our system uses the refined depths $\mathcal{D}^*$ 
to reason densely about occlusions and free space~(\cref{fig:depthcheck}).
If a registered image $I_c$ has too many conflicting observations with other registered images $I_i$, we de-register it and discard associated structure.
To decide if
an image causes inconsistencies, we reproject its depth map $D_c^*$ into overlapping other images and accumulate a min-depth buffer $D_{i\leftarrow c}^*$ with associated reprojected depth uncertainty
$\Sigma_{D_{i\leftarrow c}}$.
Vice versa, we reproject other depth maps into image $I_c$ to build $D_{c\leftarrow i}^*$ and $\Sigma_{D_{c\leftarrow i}}$.

We then compute the amount of forward-backward inconsistency as
the ratio of inconsistent pixels with confidence $\gamma$:
\begin{align}
    \beta_{i}
    &= \frac{1}{W_i H_i}\sum_{u, v} \mathds{1}\!\left(
      \frac{D_i^*(u, v) - D_{i\leftarrow c}^*(u,v)}{\Sigma_{D_i}(u, v) + \Sigma_{D_{i\leftarrow c}}(u, v)} > \gamma
    \right) \\
    & + \frac{1}{W_c H_c}\sum_{u, v} \mathds{1} \left(
      \frac{D_c^*(u, v) - D_{c\leftarrow i}^*}{\Sigma_{D_c}(u, v) + \Sigma_{D_{c\leftarrow i}}(u, v)} > \gamma
    \right)
    \enspace ,\nonumber
\end{align}
where $\mathds{1}$ is the indicator function
and $\beta_i \in [0, 2]$.
We consider a view $c$ as inconsistent if any of the overlapping views' $\beta_i$ exceeds a ratio $\hat{\beta}$ of occluded pixels.
We check the depth consistency for each newly registered view and once in the very end to discard potentially incorrect registrations.

\subsection{Implementation Details}

\paragraph{Correspondence search:} We largely rely on COLMAP's correspondence search pipeline with the main difference that we use SuperPoint~\cite{superpoint} and LightGlue~\cite{lindenberger2023lightglue} for sparse feature extraction and matching.
Since our system does not require multi-view tracks, it can handle dense correspondences as well as is -- we experiment with those of RoMA~\cite{edstedt2024roma}. MASt3R~\cite{leroy2024grounding}, on the other hand,  yields the best results when used to match sparse SuperPoint keypoints.
For scalable matching, we compute global image features~\cite{arandjelovic2016netvlad} to shortlist, per image, the $n^*$ most similar other images as candidate pairs for feature matching.

\paragraph{Monocular Depth priors:}
Our most general configuration is based on monocular depth and normals predicted by Metric3D-v2~\cite{metric3dv2}, which also estimates uncertainties. 
When matching images with MASt3R~\cite{leroy2024grounding},
we instead use its depth estimates along with normals from DSINE~\cite{bae2024dsine}.
We also experiment with other depth models \cite{depth_anything_v2,depth_pro} in \cref{sec:analysis}.

\paragraph{Refinement optimization:}
The optimization of $C_\mathrm{reg} + C_\mathrm{int}$ is implemented on the GPU, while the term $C_\mathrm{reproj} + C_\mathrm{reg}$ is solved using Ceres~\cite{ceres-solver} on the CPU.
In our experiments, we did not experience a significant difference in end-to-end accuracy metrics between using the propagated covariance $\Sigma_{D_i^*(x_j)}$ as compared to the prior covariance $\Sigma_{D_i(x_j)}$.
For faster runtimes, we therefore use $\Sigma_{D_i(x_j)}$ in our implementation of the local and global refinement optimization.

\section{Experiments}

We assess the performance of our system in both low-overlap and low-parallax scenarios.

\subsection{Low-overlap reconstruction}

\paragraph{Setup:}
We consider collections of images from several SfM datasets~\cite{schops2017multi,duckworth2023smerf,knapitsch2017tanks}.
For each scene, we sample multiple groups of images with a pre-defined maximum amount of visual overlap across views.
The overlap is computed as the ratio of covisible pixels using ground-truth depth maps, when available.
Otherwise, it is defined as the ratio of covisible feature points in the original, full SfM model.
We assume calibrated cameras.
Following standard practice~\cite{jin2021benchmark}, we evaluate the camera pose accuracy by comparing the relative poses to the ground truth.
The error is defined as the maximum of angular errors in rotation and translation.
We report the Area Under the recall Curve (AUC) up to 1/5/20\degree.

\paragraph{Approaches:}
Among those based on sparse correspondences, we consider COLMAP~\cite{schoenberger2016sfm}, with correspondences estimated by SIFT~\cite{lowe2004distinctive} and by SuperPoint~\cite{superpoint} and LightGlue~\cite{lindenberger2023lightglue}, as well as extensions for structure-less resectioning (SLR)~\cite{zheng2015structure} and global SfM with GLOMAP~\cite{pan2024glomap}.
As for dense correspondences, existing systems cannot handle them easily, because they do not form multi-view tracks.
Past works~\cite{loftrtpami,pixsfmtpami} cluster them into tracks based on proximity.
We thus run COLMAP with RoMa~\cite{edstedt2024roma}.
DF-SfM~\cite{he2024detector} combines COLMAP with LoFTR~\cite{sun2021loftr} and further refines the tracks.
VGGSfM~\cite{wang2023vggsfm} estimates correspondences for a subset of images simultaneously.
We also consider MASt3R-SfM~\cite{duisterhof2024mast3r}.
It is based on two-view correspondences and point maps, which embed some monocular priors as well.

\paragraph{Triplet evaluation:}
To minimally capture the low-overlap scenario, we first consider triplets of images, as shown in \cref{fig:teaser}.
We sample them from 25 indoor and outdoor scenes of the ETH3D dataset~\cite{schops2017multi} with different levels of three-view overlap, from zero to 50\%.
The results, in \cref{tbl:results:eth3d}-left, confirm that COLMAP fails catastrophically for triplets with the least overlap.
Learned matching (SP+LG) and structure-less resectioning both improve but are largely insufficient.
Our approach is robust (high AUC@20\degree) and is increasingly accurate with the overlap (increasing AUC@1\degree).
It significantly outperforms all existing approaches.

\paragraph{Full-scene SfM:}
We 
assemble larger 
images sets from multiple datasets.
In addition to ETH3D, we consider 7 scenes from the Tanks \& Temples dataset~\cite{knapitsch2017tanks} and 4 scenes introduced in SMERF~\cite{duckworth2023smerf}, with indoor/outdoor spaces, repeated structures, and texture-less views.
The GT camera poses were estimated with COLMAP -- achieving sufficient accuracy by using 10 to 100 times more images.
For each dataset overlap bucket, we sample 5 image collections per scene.

The results, in \cref{tbl:results:eth3d}-right and \cref{tbl:results:sfm}, show that our approach outperforms existing ones in the low-overlap scenarios, even when considering the same input correspondences.
As the overlap increases, it maintains superiority in terms of robustness, even when the views are dense (\emph{all images} on ETH3D).
Using dense features like RoMa or MASt3R works better than SuperPoint+LightGlue.
MASt3R's ability to reject negative pairs and handle extreme viewpoints yields the best overall performance.
Our approach is, however, less accurate than MASt3R-SfM at AUC@1\degree \ in the T\&T dataset, due to a lack of foreground matches in object-centric scenes.
We show qualitative examples in \cref{fig:qualitative}.

\begin{table}[t]
\centering
\scriptsize
\renewcommand{\b}[1]{\textbf{#1}}
\begin{tabular}{llccc}
\toprule
&  & \multicolumn{3}{c}{pose AUC at X\degree}\\
\cmidrule(lr){3-5}
strategy & approach & 1\degree & 10\degree & 30\degree\\
\midrule
incremental & SP+LG + COLMAP (default) & 18.3 & 49.2 & 55.1\\
incremental & SP+LG + COLMAP (tuned) & 25.8 & 71.0 & 82.4\\
incremental & SP+LG + StudioSfM & 19.9 & 72.3 & 86.5 \\
global & SP+LG + GLOMAP & 34.2 & 81.0 & 90.7 \\
incremental & SP+LG + \textbf{ours} & 30.9 & 79.5 & 90.4\\
\midrule   
global & MASt3R-SfM & 33.4 & 80.8 & 91.2\\
incremental & MASt3R + \textbf{ours} & \textbf{35.5} & \textbf{81.9} & \textbf{91.5}\\
\bottomrule
\end{tabular}
\caption{\textbf{SfM with low parallax on the RealEstate10k dataset.}
Our approach outperforms previous incremental SfM pipelines as well as MASt3R-SfM, even though global SfM does not suffer from the same fundamental low parallax limitations.
\label{tbl:results:re10k}%
}
\end{table}

\begin{table*}[t]
\centering
\scriptsize
\renewcommand{\b}[1]{\textbf{#1}}
\begin{tabular}{cllllcccc}
\toprule
\multirow{2}{*}[-0.4em]{\#} & \multicolumn{2}{c}{depth} & \multicolumn{2}{c}{surface normals} & \multicolumn{2}{c}{ETH3D dataset} & \multicolumn{2}{c}{SMERF dataset}\\
\cmidrule(lr){2-3} \cmidrule(lr){4-5} \cmidrule(lr){6-7} \cmidrule(lr){8-9}
& model & uncertainty & model & uncertainty & minimal overlap & all images & minimal overlap & high overlap\\
\midrule
1  & \textbf{Metric3D-v2}~\cite{metric3dv2} & \textbf{yes} & \textbf{Metric3D-v2}~\cite{metric3dv2} & \textbf{yes}
& 27.3/55.9/71.8 & 74.3/88.3/92.0 & 9.2/\textbf{41.0}/\textbf{69.8} & \textbf{47.3}/\textbf{79.3}/\textbf{90.6} \\

2 & Metric3D-v2~\cite{metric3dv2} & no & Metric3D-v2~\cite{metric3dv2} & yes
&  27.3/55.4/71.2 & 72.8/86.8/90.5 &  8.7/38.7/65.8 & 40.0/67.2/82.1 \\

3 & Metric3D-v2~\cite{metric3dv2} & yes & Metric3D-v2~\cite{metric3dv2} & no
& 27.3/55.8/71.7 & 74.2/88.3/91.9 &  9.1/39.9/67.2  & 43.6/76.2/89.2\\

\cmidrule(lr){4-9}
4 & Metric3D-v2~\cite{metric3dv2} & yes & DSINE~\cite{bae2024dsine} & yes
& 26.3/54.2/70.4 & 74.3/88.3/91.9 & 8.9/39.0/66.7 & 44.5/78.3/89.8 \\

5 & Metric3D-v2~\cite{metric3dv2} & no & DSINE~\cite{bae2024dsine} & yes
& 26.3/53.7/69.5 & 72.5/86.8/90.5 & 8.2/36.2/61.9 &  41.2/68.3/79.4 \\

6 & Depth Pro~\cite{depth_pro} & no & DSINE~\cite{bae2024dsine} & yes
& 23.3/47.5/63.4 & 74.5/88.3/91.9 & 7.0/30.8/57.0 &  30.7/56.5/72.7 \\

7 & Depth Anything v2~\cite{depth_anything_v2} & no & DSINE~\cite{bae2024dsine} & yes
& 21.3/44.8/60.8  & 71.8/86.2/89.9  & 4.0/19.8/43.8 & 20.7/42.2/60.4  \\

8 & M3D-v2 Large~\cite{metric3dv2} & yes & M3D-v2 Large~\cite{metric3dv2} & yes
&  \textbf{28.2}/\textbf{56.5}/\textbf{72.2} & \textbf{75.0}/\textbf{88.6}/\textbf{92.1} &  \textbf{ 9.2}/40.2/69.4 & 43.4/72.6/84.1\\

9 & M3D-v2 Small~\cite{metric3dv2} & yes & M3D-v2 Small~\cite{metric3dv2} & yes
&  26.1/54.0/70.6 & 74.3/88.0/91.8  & 6.8/33.4/62.5 & 34.5/59.8/76.4\\
\bottomrule
\end{tabular}
\caption{\textbf{Comparing monocular priors.}
Only Metric3D-v2 estimates depth uncertainties -- critical for fusing with feature correspondences 
(1 vs 2, 4 vs 5).
Models without depth uncertainties are less reliable (1 vs 6/7).
Factoring out uncertainties, Metric3D-v2 
outperforms  DepthPro and DepthAnything-v2 overall (5 vs 6 vs 7).
Having uncertainties for surface normals is also important (1 vs 3).
Both Metric3D-v2 and DSINE estimate normal uncertainties and work well with our pipeline (1 vs 4). 
While the smaller and more efficient variants of Metric3D-v2 (Large and Small) struggle on SMERF, they perform similarly as the largest model (Giant2) on ETH3D (1 vs 8/9).
\vspace{-0.5cm}
\label{tbl:results:priors}%
}
\end{table*}

\begin{table}[t]
\centering
\scriptsize
\setlength\tabcolsep{2pt}%
\renewcommand{\b}[1]{\textbf{#1}}
\begin{tabular}{lcccc}
\toprule
\multirow{2}{*}[-0.4em]{variant} & \multicolumn{2}{c}{ETH3D dataset} & \multicolumn{2}{c}{SMERF dataset}\\
\cmidrule(lr){2-3} \cmidrule(lr){4-5}
& min. overlap & all images & min. overlap & high overlap\\
\midrule
SP+LG + \textbf{ours}
& 27.3/55.9/71.8 & 74.3/88.3/92.0 & \09.2/41.0/69.8 & 47.3/79.3/90.6
\\

no depth refinement
& 26.8/55.2/69.9  & 71.9/87.6/91.7 & \08.4/37.6/66.7 & 32.2/63.6/82.0
\\
no depth reg.
& 23.6/49.6/65.9 & 75.1/88.7/92.2 & \05.7/21.0/45.9 & 45.5/64.1/73.0
\\
ground-truth depth
& 42.9/68.0/77.1 & 73.8/87.3/90.8 & - & -
\\
no lifting
& 10.6/16.1/18.7 & 74.1/87.2/90.6 & \01.0/\01.7/\02.1 & 51.9/69.6/75.2   
\\

\midrule
ROMA + \textbf{ours}
& 33.4/60.6/74.4 & 71.6/87.0/91.3 & 10.6/41.0/61.8 & 41.4/69.3/79.4
\\

no depth refinement
& 30.9/60.1/74.5  & 66.6/85.3/90.8 & \08.8/35.7/57.4 & 29.3/61.1/77.5
\\
no depth reg.
& 29.1/50.5/65.3 & 72.1/87.6/91.8  & \07.2/24.8/45.4 & 54.0/73.6/80.6
\\
no lifting
& 13.5/19.2/22.4 & 70.4/84.6/88.1   &  \01.4/\02.3/\02.9  & 55.0/72.9/77.8 
\\ 
\bottomrule
\end{tabular}
\caption{\textbf{Ablation of our pipeline.}
Refining the prior depth with surface normals is consistently effective as they are complementary.
The depth regularization is useful in low overlap.
It can impair the accuracy for under-constrained 2-view tracks
because monocular priors provide noisier constraints than reprojection errors.
Using the ground-truth LiDAR depth, which is sparser but not biased, largely improves the accuracy at 1\degree.
While lifting greatly improves low overlap reconstruction, it introduces noise in well-posed scenarios.
\label{tbl:results:ablation}%
}
\end{table}

\subsection{Low-parallax reconstruction}
As incremental SfM relies on noisy 3D structure for registration, it is typically prone to failure in low-parallax scenarios.
\paragraph{Setup:}
Following past research~\cite{wang2023posediffusion,duisterhof2024mast3r}, we consider the RealEstate10k dataset~\cite{zhou2018stereo}.
It features indoor and outdoor scenes with low texture and challenging camera motion, \eg, forward translation and in-place rotation.
We evaluate COLMAP and GLOMAP~\cite{pan2024glomap} with SuperPoint and Lightglue~\cite{superpoint,lindenberger2023lightglue} and MASt3R-SfM~\cite{duisterhof2024mast3r}.
We found that relaxing the minimum triangulation angle allowed in COLMAP is critical.
We report results for this variant in addition to one using the default parameters.
We also evaluate StudioSfM~\cite{studiosfm}, which is designed for this low-parallax scenario.
It also leverages monocular depth but only for initialization, 
and regularization, and it does not refine it or handle uncertainties.

\paragraph{Results:} 
In \Cref{tbl:results:re10k}, we distinguish between global and incremental SfM, as global methods do not rely on noisy 3D structures for registration.
MP-SfM largely closes the gap between incremental and global SfM and outperforms COLMAP, StudioSfM and MASt3R-SfM, likely because they do not handle well uncertainties of monocular priors.

\subsection{Ablation Study}
\label{sec:analysis}
We present ablations to justify the design of our system.

\paragraph{Impact of the monocular priors:}
In \cref{tbl:results:priors}, we compare the SfM accuracy with depth and normals estimated by different approaches and we study the impact of the prior uncertainty.

\paragraph{Impact of each component:}
In \cref{tbl:results:ablation}, we evaluate the impact of the depth refinement, regularization, and lifting.
All components are beneficial in different scenarios.

\paragraph{Robustness to symmetry:}
In \cref{tbl:results:depth-check}, we show that our depth consistency check is effective in scenes with symmetries.
We compare it to Doppelgangers~\cite{cai2023doppelgangers} as an approach that also rejects non-matching image pairs but based on a neural network.
We use it to pre-filter pairs or to rerank images in the incremental process.

\begin{table}[t]
\centering
\scriptsize
\setlength\tabcolsep{1.6pt}%
\renewcommand{\b}[1]{\textbf{#1}}
\begin{tabular}{lcccc}
\toprule
\multirow{2}{*}[-0.4em]{variant} 
& \multicolumn{2}{c}{ETH3D dataset} 
& \multicolumn{2}{c}{SMERF dataset}\\
\cmidrule(lr){2-3}
\cmidrule(lr){4-5}
& min. overlap & all images 
& min. overlap & high overlap\\
\midrule
no check 
& 26.9/55.7/71.5 & 74.7/88.4/92.0 %
& 8.7/39.1/66.9  &  32.8/58.8/72.8\\   
Doppelgangers (filter) 
& 21.7/39.6/50.0 & 70.2/83.5/86.9 %
& 0.2/1.0/1.4 & 8.3/15.9/22.6\\
Doppelgangers (rank) 
& 26.9/\textbf{56.0}/\textbf{72.1} & \textbf{77.0}/\textbf{91.4}/\textbf{94.8} %
& 8.5/38.9/66.5  & 32.9/57.0/67.5\\   
\textbf{our check} 
& \textbf{27.3}/55.9/71.8  & 74.3/88.3/92.0 %
&  \textbf{9.2}/\textbf{41.0}/\textbf{69.8}  &\textbf{ 47.3}/\textbf{79.3}/\textbf{90.6}\\ 
\bottomrule
\end{tabular}
\caption{\textbf{Evaluation of the depth consistency check.}
Our check significantly improves the camera poses on scenes with repeated structures (SMERF).
It is more effective than the learning-based approach Doppelganger, which is affected by domain shift.
\label{tbl:results:depth-check}%
}
\end{table}

\paragraph{Efficiency:}
Extracting depth and normals takes \qty{0.65}{\second} per image. Sky masks are also extracted to handle overly confident depth estimates, adding \qty{0.12}{\second} per image.
Refining the depth takes on average \qty{70}{\milli\second} per image, is done for only few images, and is faster as the poses and 3D points are less updated throughout the incremental process. 
The depth check takes \qty{14}{\milli\second} per newly registered image.

\section{Limitations}
Our system depends on reliable uncertainties for the monocular priors. 
State-of-the-art depth models rarely estimate uncertainties and those that do are often over-confident.
Estimating accurate normals is also difficult for some types of surfaces, such as vegetation, limiting the  performance outdoors.

Furthermore, while our approach pushes SfM beyond previously tackled scenarios, it still relies on standard components that often fail in the harsh environments we address.

For example, image retrieval struggles with strong perspective changes and symmetries, while image matching often cannot handle extremely low visual overlap.
We explore solutions to these limitations, such as the depth consistency check, but our approach will also seamlessly benefit from future improvements with little adaptation.
Lastly, our approach increases the run time of SfM, which is often only the first step of complex applications.

\section{Conclusion}
This paper introduces an approach that significantly improves the robustness of SfM in low-overlap, low-parallax and high symmetry scenarios, which are common failure cases in data captured by non-expert users.
To achieve this, we augment the classical SfM paradigm with monocular depth and surface normals estimated by deep neural networks.
Our system leverages these priors in multiple ways: at initialization, when registering new views, when optimizing the poses and geometry, and to reject incorrect registrations, thereby making SfM more robust to repeated patterns.
This results in improvements across different image correspondences, environments, and levels of difficulty.

\begin{figure*}[tp]
    \centering
    \setlength{\pheight}{-1.5mm}
    \setlength{\pwidth}{0.005\linewidth}
    \setlength{\iwidth}{0.312\linewidth}
    \setlength{\lwidth}{0.312\linewidth}

    \includegraphics[width=\iwidth]{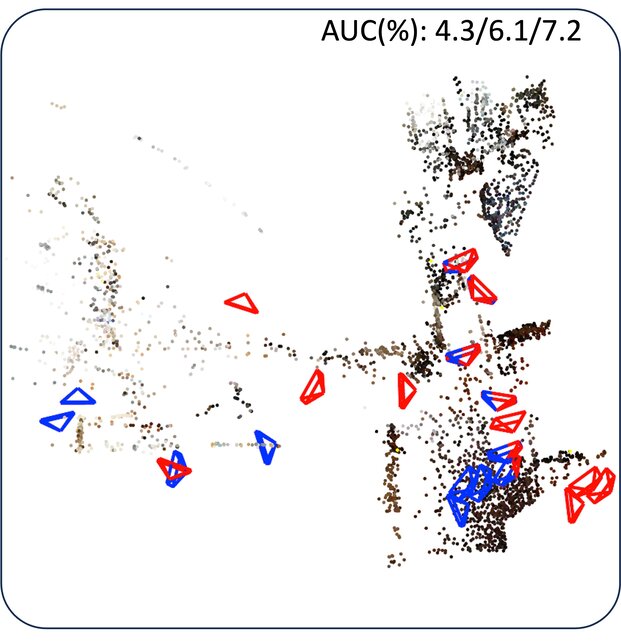}%
    \hspace{\pwidth}%
    \includegraphics[width=\lwidth]{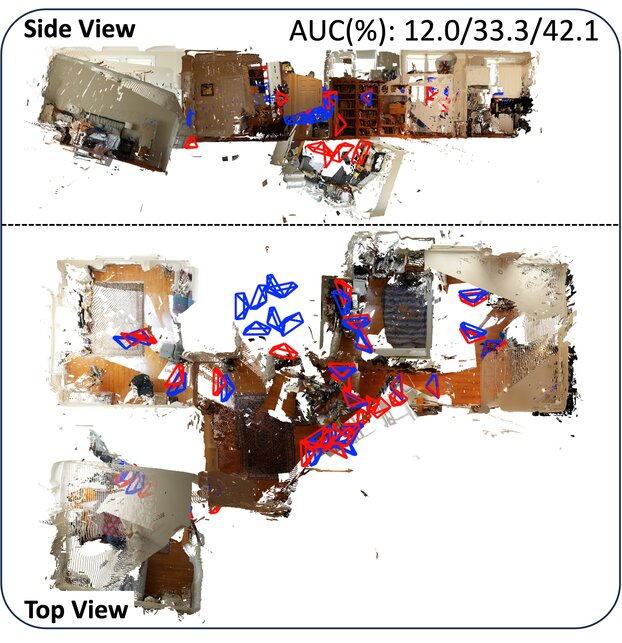}
    \hspace{0.0\linewidth}%
    \includegraphics[width=\lwidth]{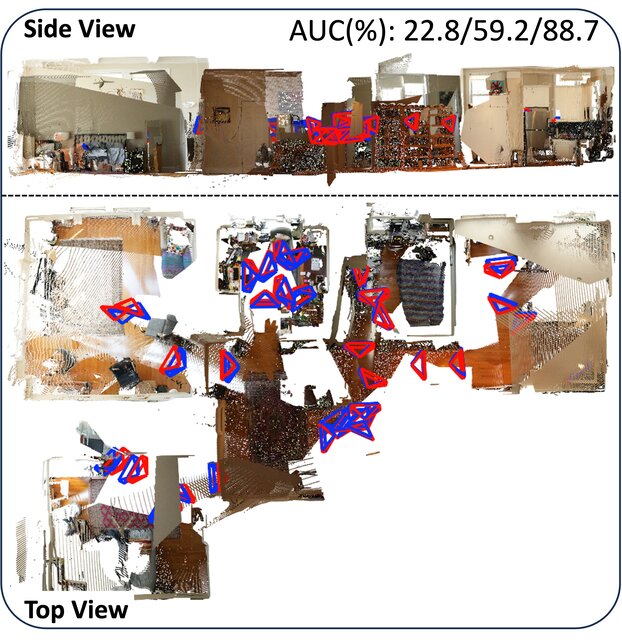}%
    \vspace{\pheight}
    
    \includegraphics[width=\iwidth]{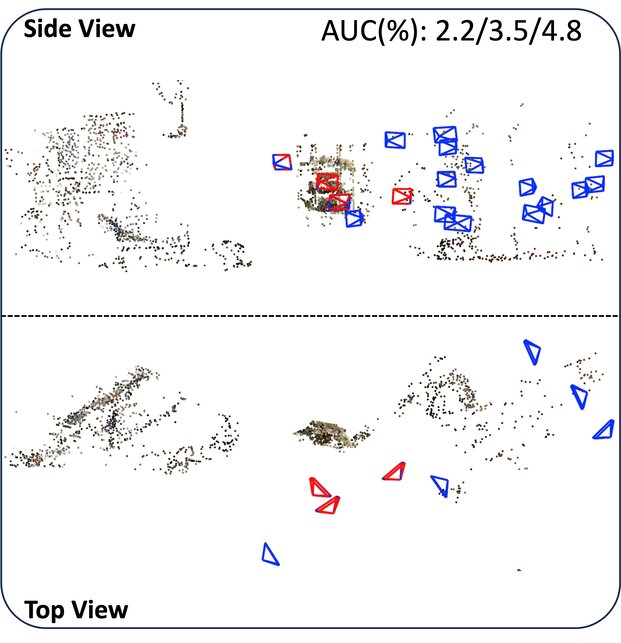}
    \hspace{0.0\linewidth}%
    \includegraphics[width=\lwidth]{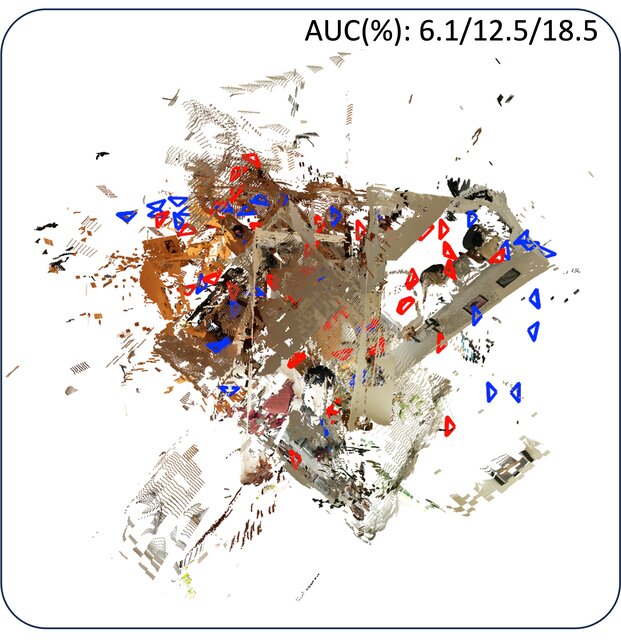}
    \hspace{0.0\linewidth}%
    \includegraphics[width=\lwidth]{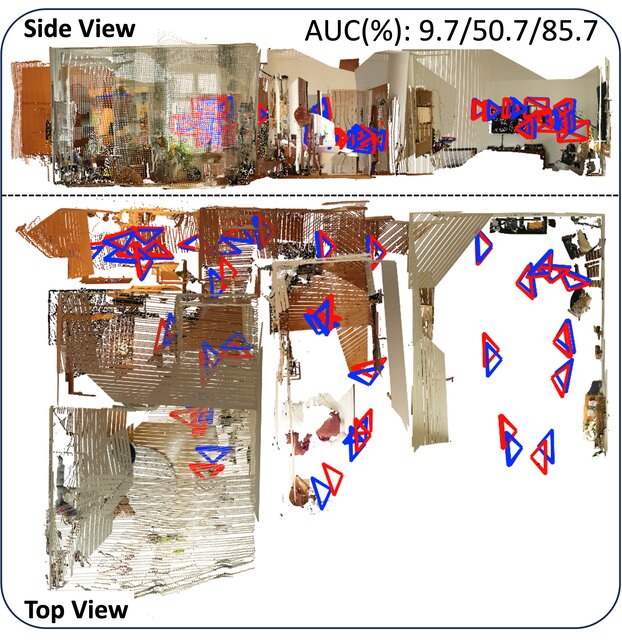}%
    \vspace{\pheight}
    
    \includegraphics[width=\iwidth]{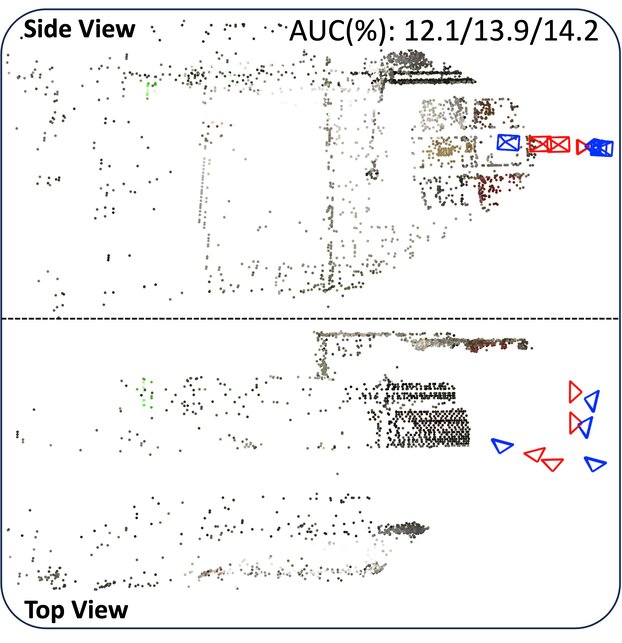}%
    \hspace{\pwidth}%
    \includegraphics[width=\lwidth]{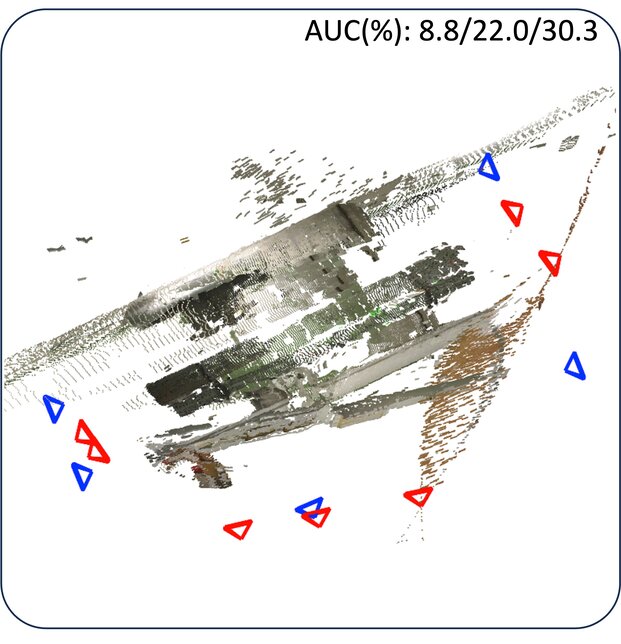}%
    \hspace{\pwidth}%
    \includegraphics[width=\lwidth]{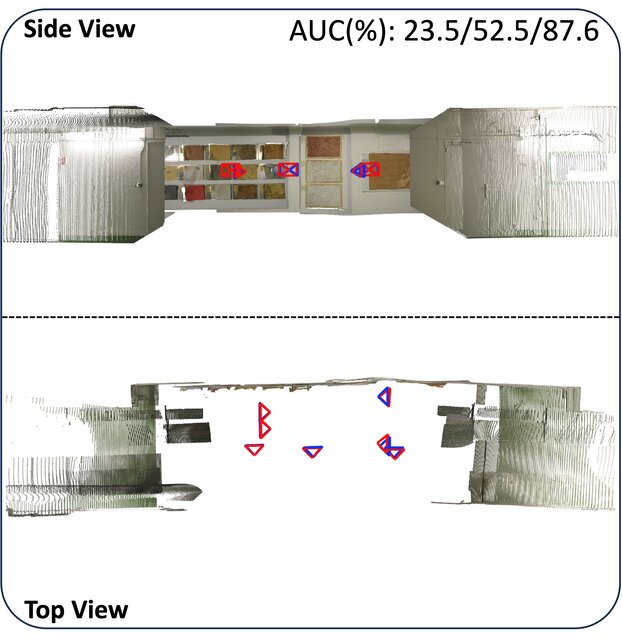}%
    \vspace{\pheight}

    \includegraphics[width=\iwidth]{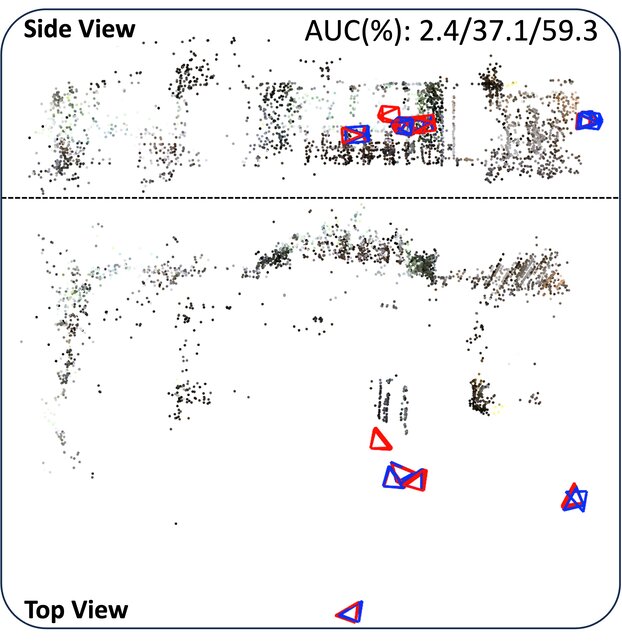}%
    \hspace{\pwidth}%
    \includegraphics[width=\iwidth]{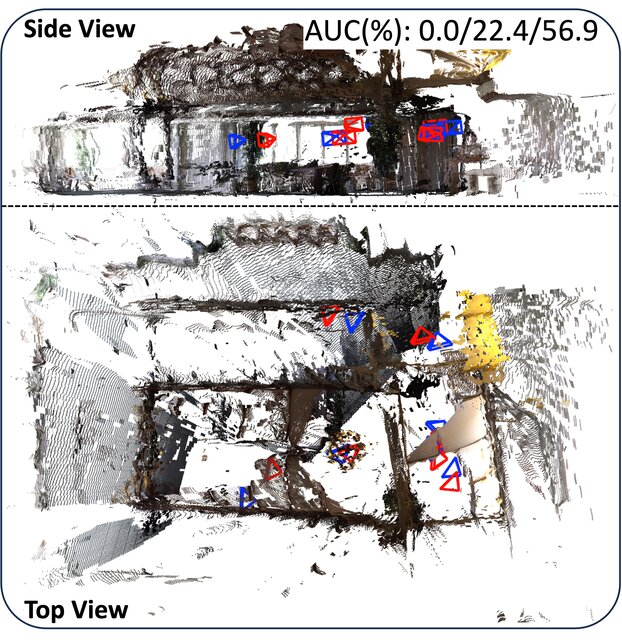}%
    \hspace{\pwidth}%
    \includegraphics[width=\iwidth]{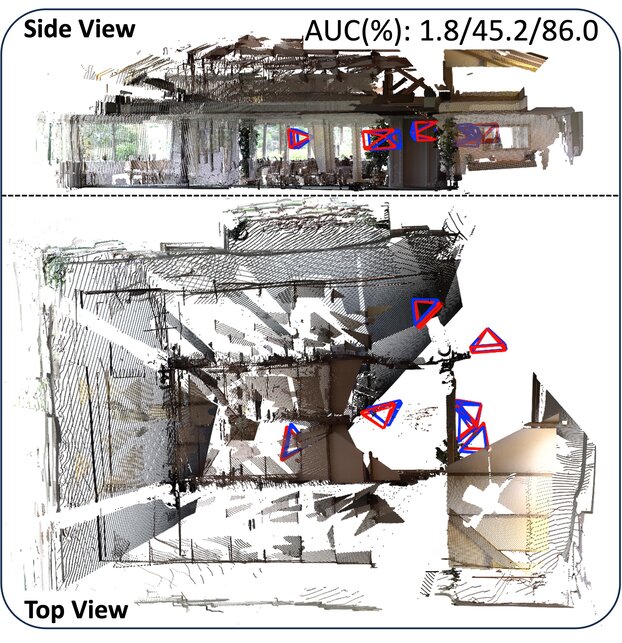}%
    \vspace{\pheight}

    \caption{\textbf{Qualitative comparison of reconstructions for low-overlap scenes.} \red{Estimated (red)} and \blue{ground-truth (blue)} camera poses, and AUC accuracies at $1^\circ/5^\circ/20^\circ$ error thresholds are presented. Left: COLMAP \cite{schoenberger2016sfm}. Center: MASt3R-SfM \cite{duisterhof2024mast3r}. Right: Our method. Rows 1–2 show scenes from SMERF~\cite{duckworth2023smerf}, while rows 3–4 are from ETH3D~\cite{schops2017multi} and Tanks and Temples~\cite{knapitsch2017tanks}.}

    \label{fig:qualitative_appendix}%
\end{figure*}

\section*{Acknowledgements} 
The authors thank Shaohui Liu for useful discussions. This work was supported by the Swiss National Science  Foundation Advanced Grant 216260: Beyond Frozen Worlds: Capturing Functional 3D Digital Twins from the Real World.

\ifproceedings
\else
    \ifaddappendix
        \appendix

\ifproceedings
\pagestyle{plain}
\begin{strip}
    \centering
    {\Large \bf MP-SfM: Monocular Surface Priors for Robust Structure-from-Motion\par}
    \vspace{0.5cm}
    {
        \large
        CVPR 2025
        \\
        \vspace{0.3cm}
        \textbf{Supplementary material}
    }
\end{strip}
\setcounter{page}{1}
In the following pages, we present additional details on the experiments conducted in the main paper.
\else
\section*{Supplementary material}
\fi

\section{Qualitative results}

In \cref{fig:qualitative_appendix}, we show reconstruction results of COLMAP \cite{schoenberger2016sfm}, MASt3R-SfM \cite{duisterhof2024mast3r}, and our approach in low-overlap scenarios.
COLMAP, which relies on three-view tracks, fails to register many of the images.
MASt3R-SfM successfully registers all images but does so in an incorrect way.
It struggles in larger scenes (rows 1 and 2) and in the presence of symmetries (rows 3 and 4).
In contrast, our pipeline successfully reconstructs these scenes with high accuracy.

\section{Depth refinement}

To refine our depth maps, we use the normal integration approach introduced by Cao~\etal~\cite{cao2022bilateral}.
We extend their cost function to account for uncertainties in the monocular surface normals.
The residual of the normal component of our integration cost is $\*r_i(u,v)=N_i(u, v) - \Delta D^*_i(u, v) \in \real^4$ for each image $i$ and pixel $(u,v)$.
We drop the indexing for the remainder of this section.
The residual is expressed as
\begin{equation}
\begin{aligned}
    \*r = \begin{pmatrix}
        r_u^+\\
        r_u^-\\
        r_v^+\\
        r_v^-
    \end{pmatrix} = \begin{pmatrix}
        \tilde{n}_{z,x}\ \partial_u^+ D^* + n_x \\
        \tilde{n}_{z,x}\ \partial_u^- D^* + n_x \\ 
        \tilde{n}_{z,y}\ \partial_v^+ D^* + n_y \\ 
        \tilde{n}_{z,y}\ \partial_v^- D^* + n_y
    \end{pmatrix}
    \enspace,
\end{aligned}
\end{equation}
where the terms 
\begin{equation}
    \begin{aligned}
        \tilde{n}_{z,x} &= n_x (u-c_x) + n_y (v-c_y) + n_z \cdot f_x, \\
        \tilde{n}_{z,y} &= n_x (u-c_x) + n_y (v-c_y) + n_z \cdot f_y, \\
    \end{aligned}
\end{equation} 
simplify the perspective case of the normal integration equations.
$n_x$, $n_y$ and $n_z$ are the three component of each normal estimate, and $f$ and $c \in \real^2$ are the focal length and principal point of the camera, while $\partial_{u/v}^{\pm} D^*$ are the discretized one-sided partial derivatives of the refined depth map $D^*$~\cite{cao2022bilateral}. 

To minimize the integration cost jointly with other costs, they should be weighted by their uncertainties.
As such, we propagate the normal uncertainties $\Sigma_N$ into residual uncertainties 
$\Sigma_{\*r} = \mathrm{diag}(\sigma_{N_u^+}^2, \sigma_{N_u^-}^2, \sigma_{N_v^+}^2, \sigma_{N_u^-}^2)$.
In the following, we derive and approximate $\sigma_{N_{u/v}^{\pm}} \approx \sigma_{N_{u/v}} $.

Monocular normal estimators \cite{metric3dv2, bae2024dsine} estimate angular isotropic uncertainties, which are projected from the Spherical into the Cartesian coordinate system using 
\begin{equation}
    \Sigma_{xyz} = J_{xyz}\,\Sigma_{\theta, \varphi}\,J_{xyz}^\top, 
\end{equation}
where
\begin{equation}
\begin{aligned}
    \Sigma_{\theta\varphi} &= 
    \begin{bmatrix}
        \sigma_\theta^2 & 0\\
        0 & \sigma_\varphi^2
    \end{bmatrix}\\
     J_{xyz} &= 
    \begin{bmatrix}
        \cos\theta\cos\varphi & -\sin\theta \sin\varphi \\
        \cos\theta \sin\varphi & \sin\theta \cos\varphi \\
        -\sin\phi & 0
    \end{bmatrix}
\enspace.
\end{aligned}
\end{equation}

Then, we express the uncertainties in the residual space $\*{r}$ as
\begin{equation}
    \sigma_{N_{u/v}}^2 = J_{u/v}\,\Sigma_{xyz}\,J_{u/v}^\top
    \enspace,
\end{equation}
where 
\begin{equation}
\begin{aligned} 
    J_u &= \begin{bmatrix}
(u - c_u) \partial_u D^* + 1 & (v - c_v) \partial_u D^* & \partial_u D^* \, f_x 
\end{bmatrix},\\
J_v &= \begin{bmatrix}
(u - c_u) \partial_v D^* & (v - c_v) 
\partial_v D^* + 1
& \partial_v D^* \, f_y \\
\end{bmatrix}.
\end{aligned}
\end{equation}
Here, to make the computation tractable, we approximate $\partial_u z = -\frac{n_x}{\tilde{n}_z} \approx\partial_u^{\pm} z$ and $\partial_u z = -\frac{n_y}{\tilde{n}_z} \approx\partial_v^{\pm} z$.

In addition to the normal uncertainty estimates, we approximate normal uncertainties using a flip consistency check between the normal estimates of the original and flipped images. In the spherical coordinates, we compute their mean $\bar{n}_{\theta,\varphi}$ and covariance
\begin{equation}
    \Sigma_{\theta, \varphi} = \begin{bmatrix}
        \angle_{1,\theta}^2 + \angle_{2,\theta}^2 & \angle_{1,\theta}\angle_{1,\varphi} +  \angle_{2,\theta}\angle_{2,\varphi} \\
        \angle_{1,\theta}\angle_{1,\varphi} +  \angle_{2,\theta}\angle_{2,\varphi} & \angle_{1,\varphi}^2 + \angle_{2,\varphi}^2,
    \end{bmatrix}
\end{equation}
where $\angle_{1,\theta}$ is the angular difference between the $\theta$ component of the original image $n_{1,\theta}$, and $\bar{n}_\theta$. As we theorize that it is better to overestimate the uncertainties, we take the maxima between the estimate uncertainties and the uncertainties derived from flip consistency.

\Cref{fig:priorvsrefined} compares monocular depth priors and refined depth maps to the ground truth. In contrast to the priors, our refined depth maps are aligned with the ground truth. 
\begin{figure*}[p]
    \centering
    \setlength{\pheight}{5mm}
    \setlength{\pwidth}{0.005\linewidth}
    \setlength{\iwidth}{0.29\linewidth}
    \setlength{\lwidth}{\dimexpr(0.999\linewidth - 2\pwidth - \iwidth)/2 \relax}

    \includegraphics[width=\iwidth]{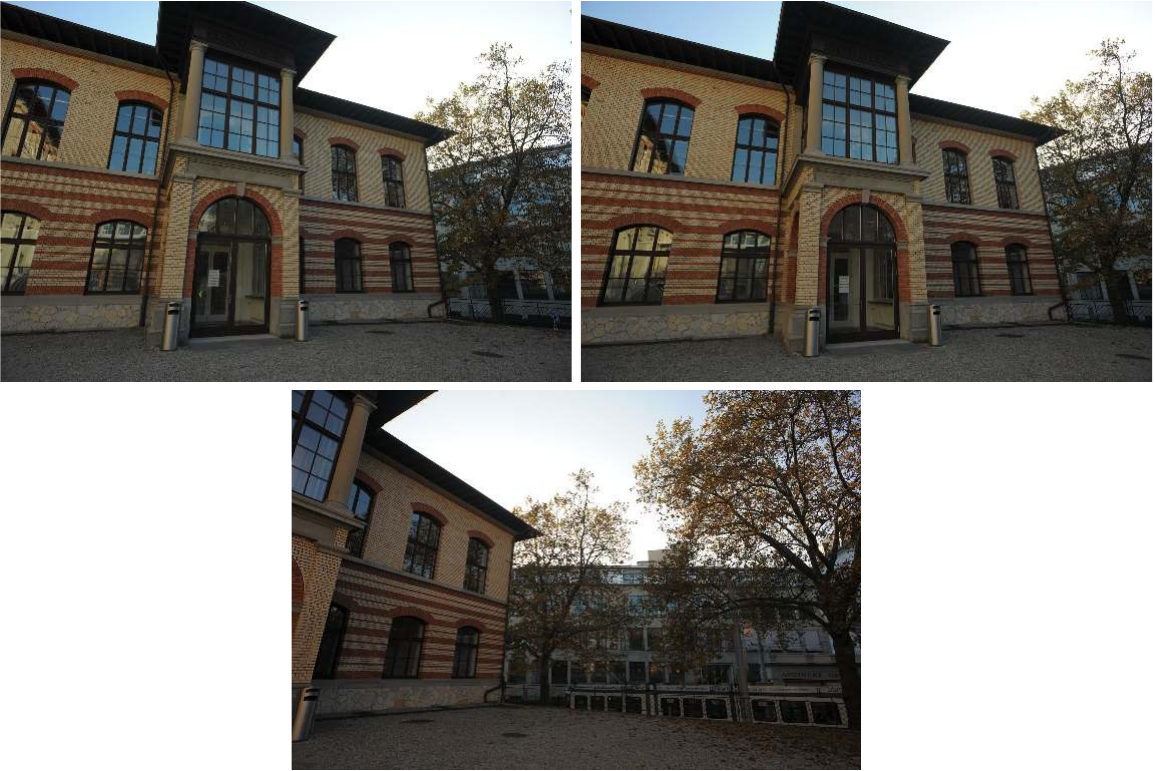}%
    \hspace{\pwidth}%
    \includegraphics[width=\lwidth]{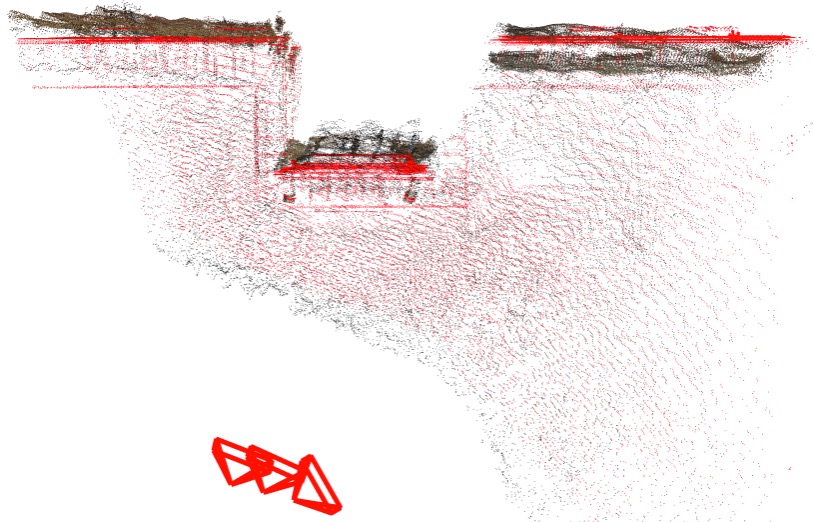}%
    \hspace{\pwidth}%
    \includegraphics[width=\lwidth]{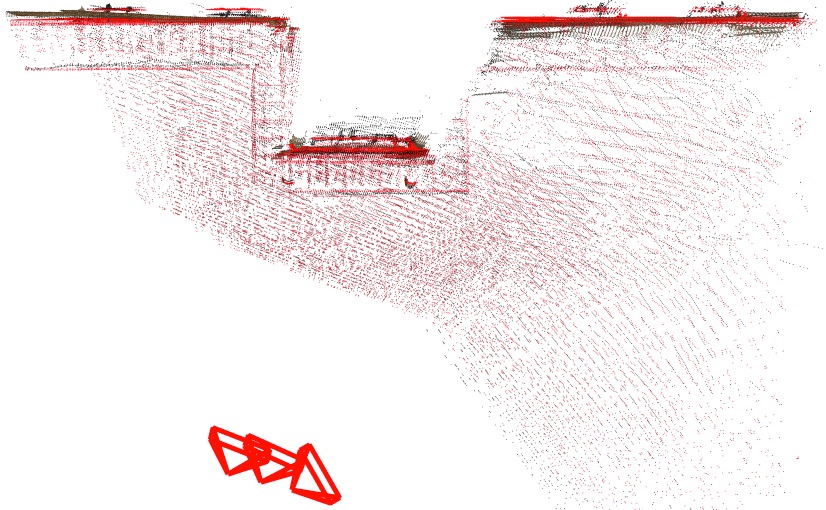}%
    \vspace{\pheight}
    
    \includegraphics[width=\iwidth]{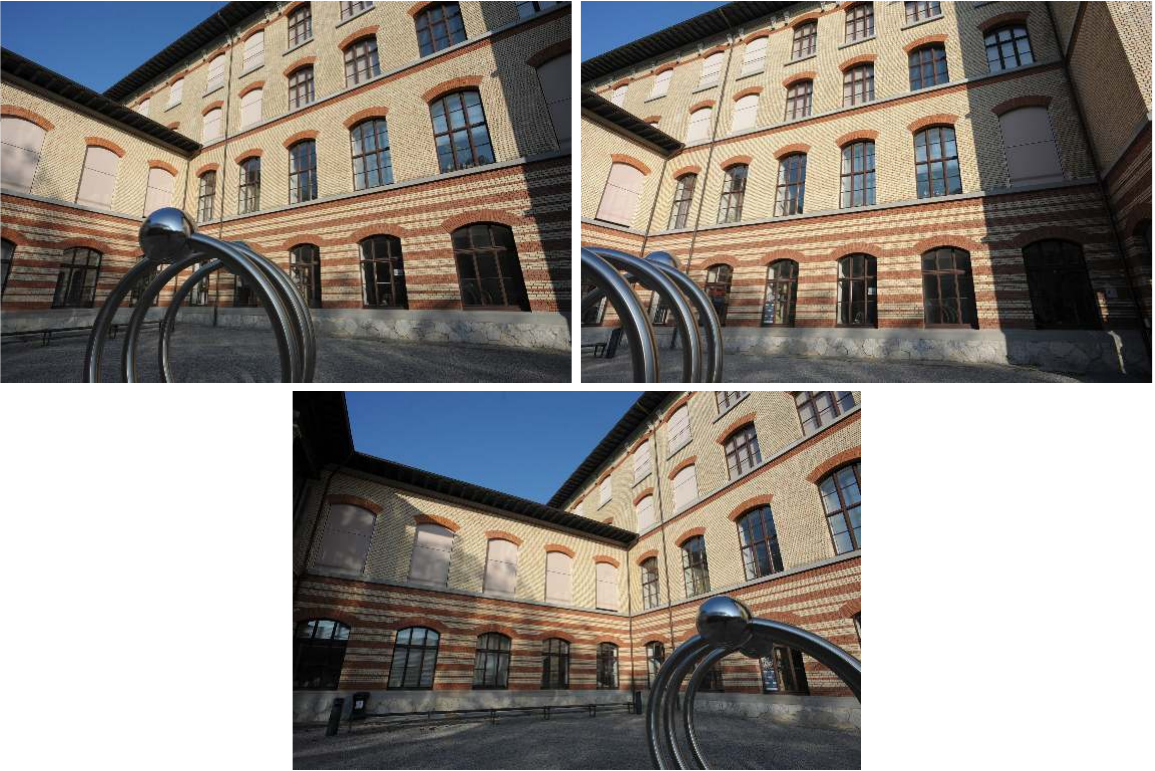}%
    \hspace{\pwidth}%
    \includegraphics[width=\lwidth]{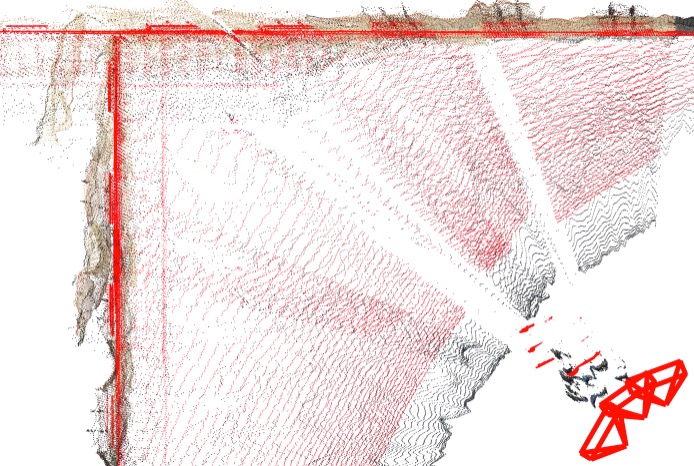}%
    \hspace{\pwidth}%
    \includegraphics[width=\lwidth]{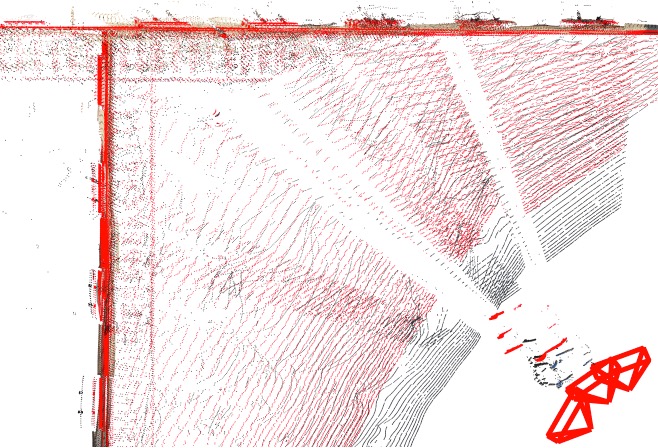}%
    \vspace{\pheight}

    \includegraphics[width=\iwidth]{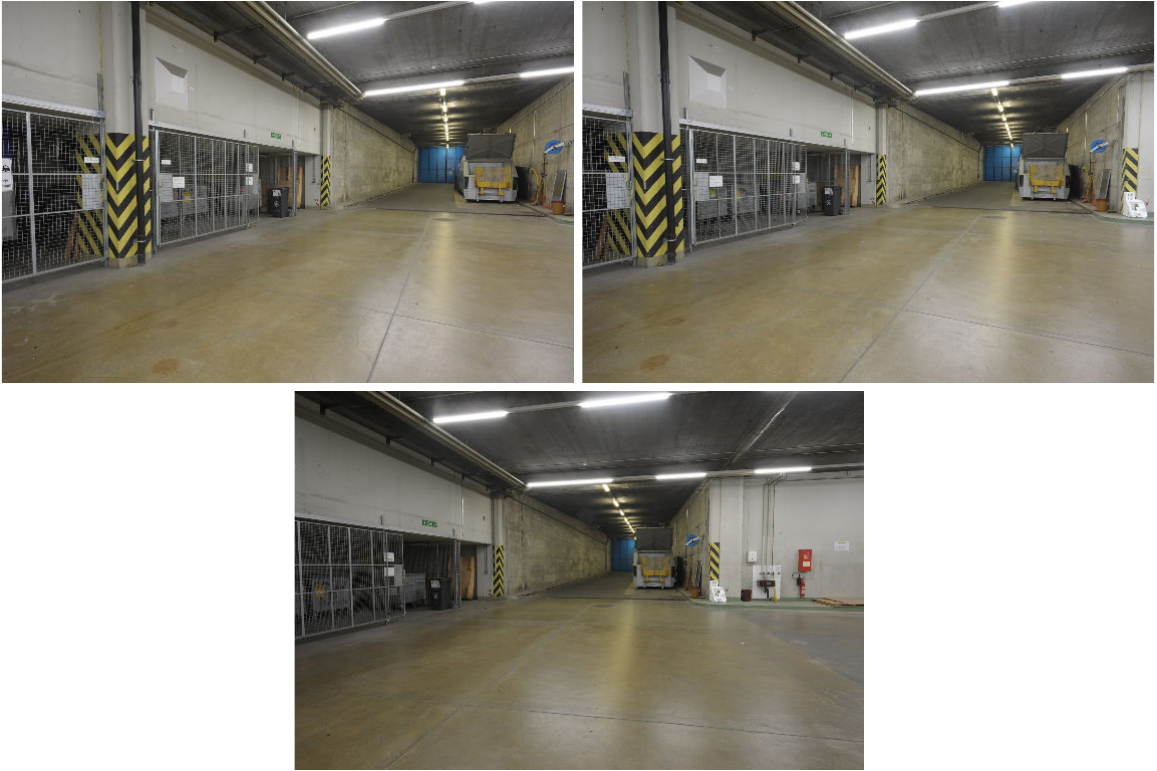}%
    \hspace{\pwidth}%
    \includegraphics[width=\lwidth]{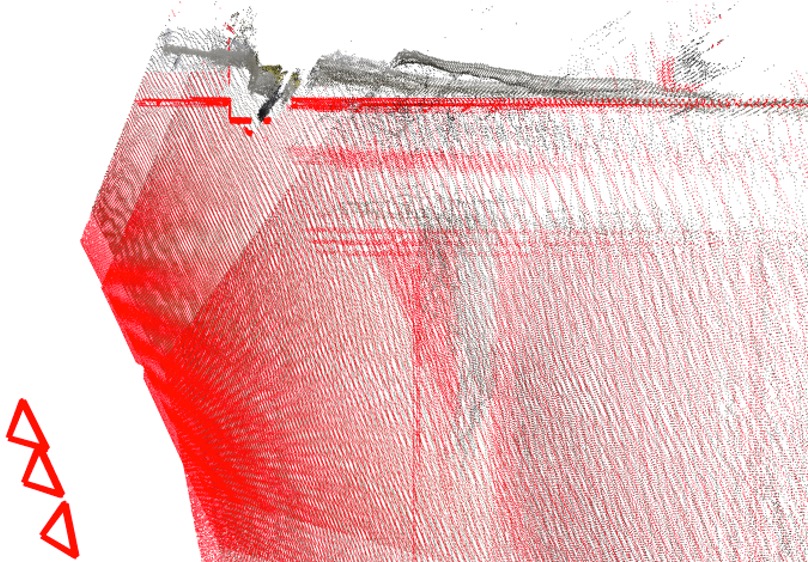}%
    \hspace{\pwidth}%
    \includegraphics[width=\lwidth]{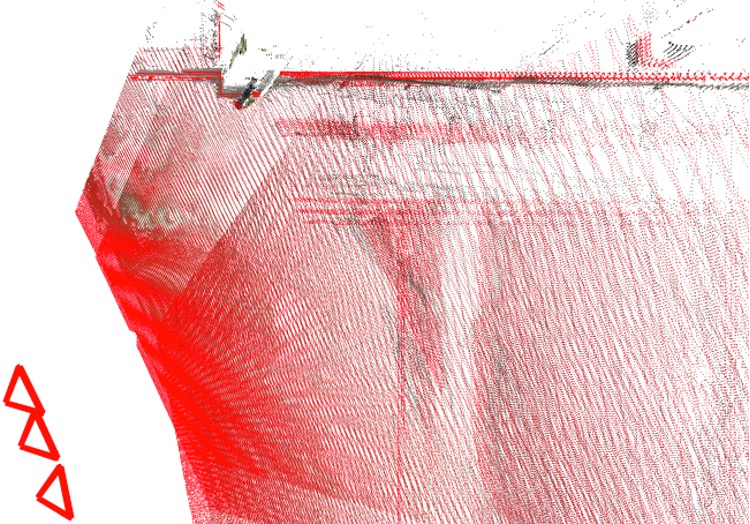}%
    \vspace{\pheight}
    
    \includegraphics[width=\iwidth]{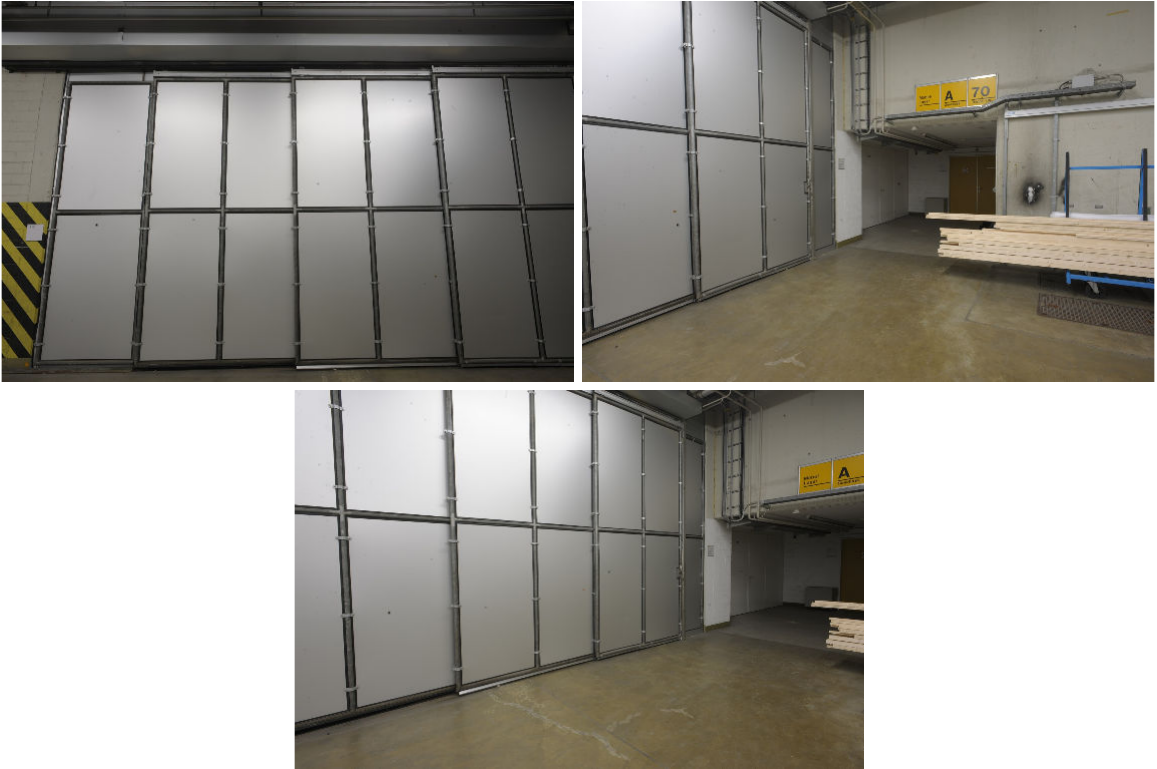}%
    \hspace{\pwidth}%
    \includegraphics[width=\lwidth]{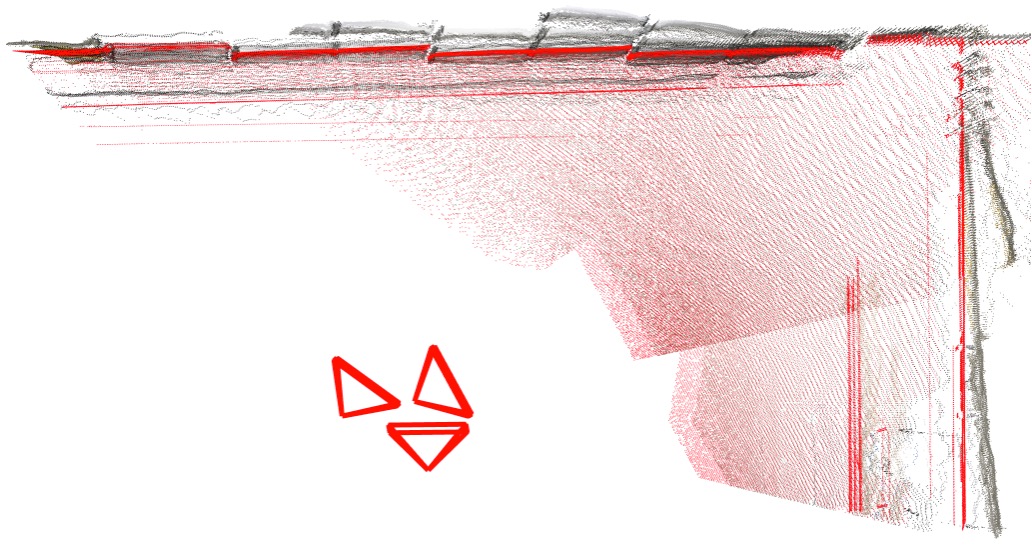}%
    \hspace{\pwidth}%
    \includegraphics[width=\lwidth]{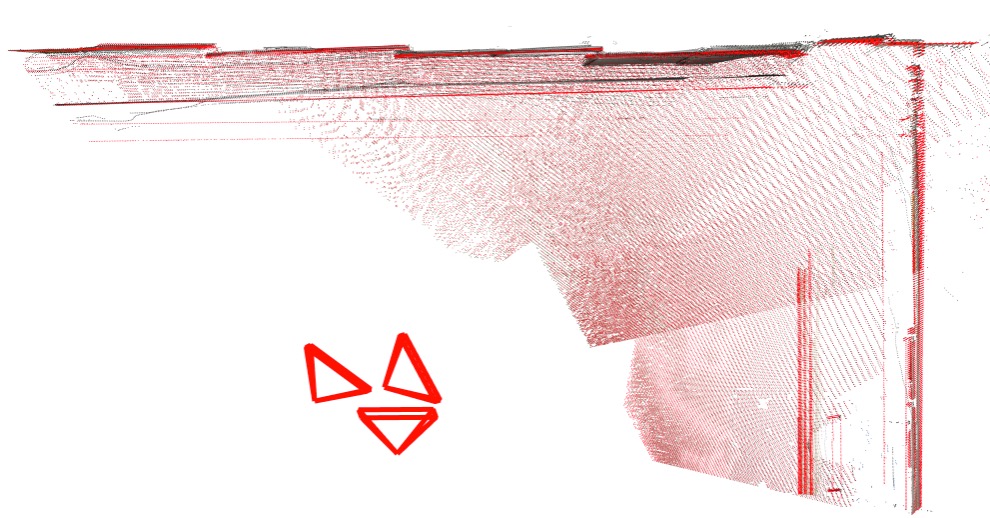}%
    \vspace{\pheight}

    \caption{\textbf{Visualizations of prior and refined depth map.}
    For four reconstructions of the ETH3D datasets, we show the input images (left) and the colored point clouds obtained by unprojecting the monocular prior depth maps (center) and the refined depth maps (right).
    We overlay the points obtained using the \red{ground-truth maps in red}.
    The refined depth maps are closer to the ground truth and more consistent across views.
    }%
    \label{fig:priorvsrefined}%
\end{figure*}

\begin{figure*}[t]
    \centering
    \includegraphics[width=0.95\linewidth]{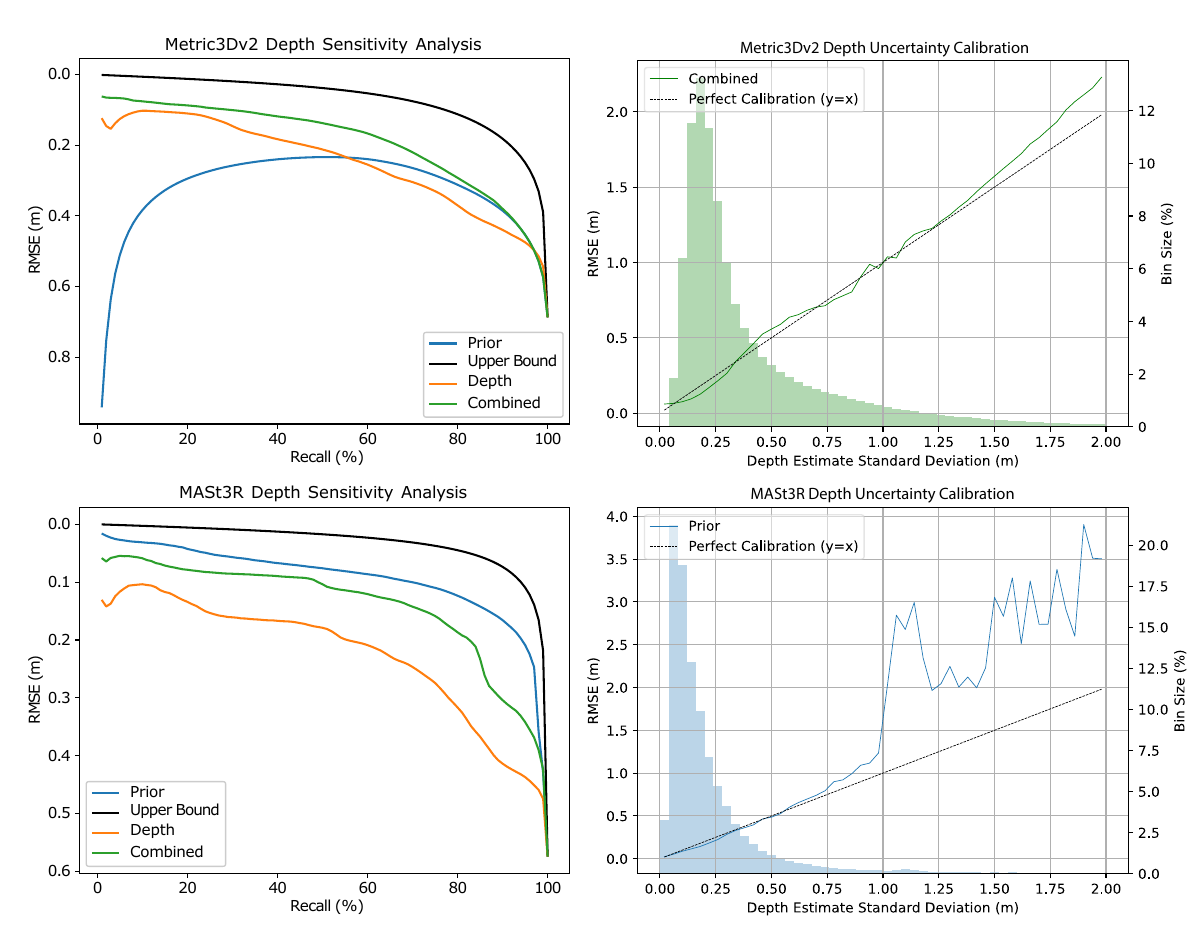} 
    \caption{\textbf{Analysis of the prior depth uncertainties on the ETH3D \cite{schops2017multi} dataset.} 
    Left: The sensitivity analysis of Metric3D-v2~\cite{metric3dv2} and MASt3R~\cite{duisterhof2024mast3r} depth estiamtes shows the total RMSE for the X\% of pixels (recall) with the lowest uncertainty. The uncertainties include the \textcolor[RGB]{31,119,180}{monocular prior} prediction, a \textcolor[RGB]{255,127,14}{depth proportional uncertainty}, and the \textcolor[RGB]{44,160,44}{combination} of the two, combined via per pixel maxima. The \emph{Upper Bound} is based on ground truth RMSE. While we found that using the combination yileded more reliable uncertianties for Metric3D-v2, the prior uncertainties predicted by MASt3R alone was the most reliable. Right: The calibration plot for the selected combination of depth uncertainties per model, optimized using a constant scaling factor. Histograms show the amount of depth estimates belonging to the estimated uncertainty bins.
    }

    \label{fig:depth_uncertainty}%
\end{figure*}

\section{Prior uncertainties}
We analyze the calibration of the uncertainties predicted by Metric3D-v2~\cite{metric3dv2} and MASt3R~\cite{leroy2024grounding} for the depth priors.
We consider the training split of the ETH3D~\cite{schops2017multi}, which has sparse ground truth depth maps obtained with laser scanners.
\Cref{fig:depth_uncertainty} shows calibration plots that compare each uncertainty with the actual depth error. 
We calibrate the uncertainties by scaling them down by a constant factor which was tuned on a different dataset given sparse SfM point clouds as pseudo ground truth.
This improves the calibration overall, except for the 10\% most confident pixels in Metric3D-v2 depth estimates.

To handle these unreliable uncertainties, we clip the standard deviations at a minimum of \qty{2}{\centi\meter} and we augment them with an uncertainty proportional to the depth estimate. 
The total uncertainty is the maximum of the scaled uncertainty and the depth-proportional uncertainty. We select the scaling factor of the depth-proportional uncertainty by maximizing the AUC of the sensitivity analysis plots (left).
We also apply robust loss functions during bundle adjustment and depth refinement.

In the case of MASt3R depth, the predicted depth uncertainty alone was the most reliable. To calibrate the other monocular depth estimators explored in \cref{tbl:results:priors}, we followed a similar approach. Although we used the same setup for the different sizes of the Metric3D-v2 models, Depth Anything V2~\cite{depth_anything_v2} and Depth Pro~\cite{depth_pro} do not predict uncertainties. We explored using a flip consistency check to estimate uncertainties; however, the improvements over the depth-proportional uncertainties were marginal.

\section{Leveraging two view correspondences}
\begin{figure*}[t]
    \centering
    \includegraphics[width=\linewidth]{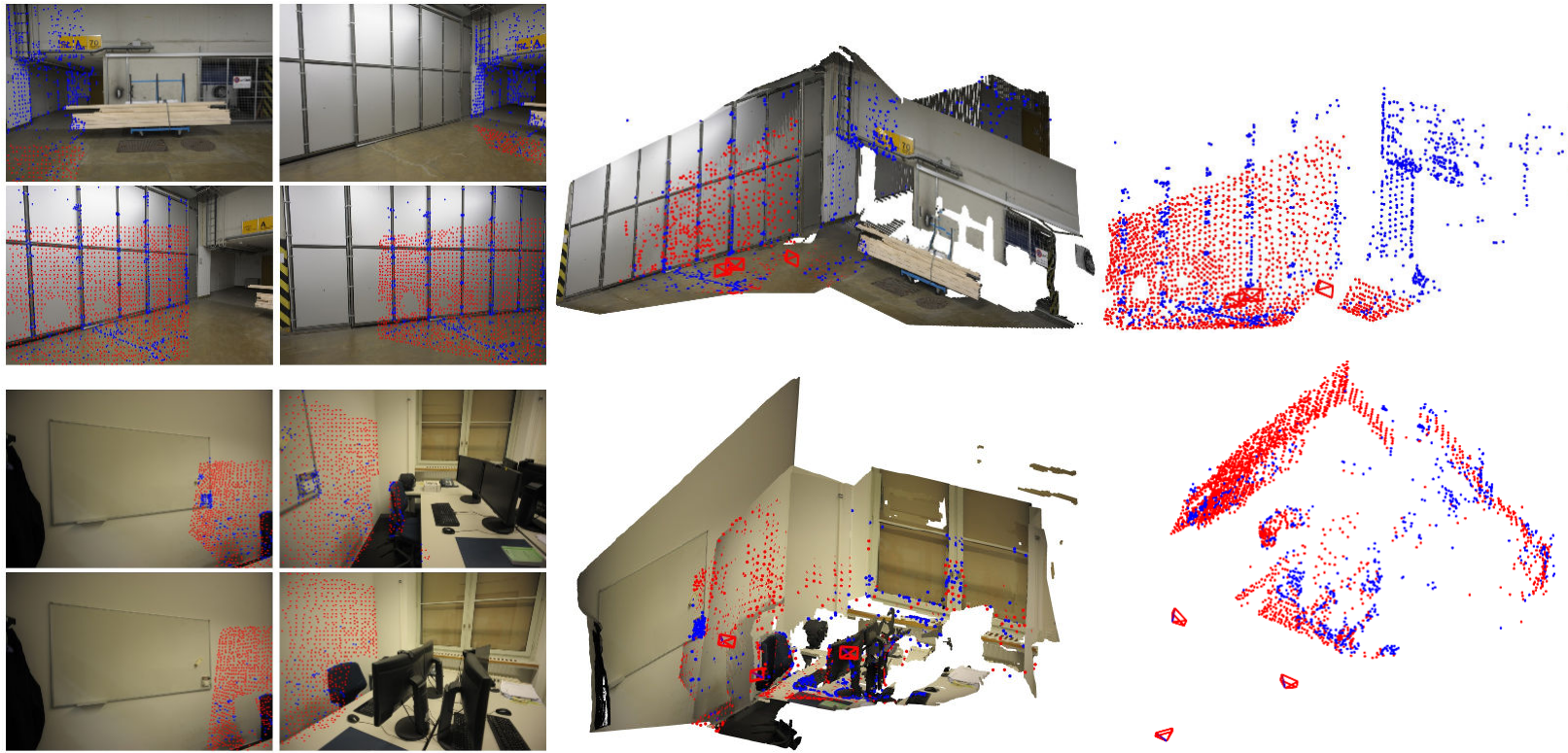} 
    \caption{\textbf{Leveraging dense matching to build long tracks in salient regions and two-view tracks in featureless areas.} Left: sparse matches sampled at {\color{blue}SuperPoint~\cite{superpoint} keypoints}, and in {\color{red}textureless areas}. Middle: dense reconstruciotn of the scene with the colmap points overlayed. Right: Visualization of only the triangulated points.
    }

    \label{fig:sparse_dense}%
\end{figure*}
  
In \cref{fig:sparse_dense}, we illustrate our approach for leveraging dense matches in non-salient regions. Building long tracks in salient regions is crucial for high accuracy in high-overlap scenarios. However, in low-overlap scenarios, leveraging matches in non-salient regions improves performance during both registration and bundle adjustment. For this reason, we use dense matchers not only to match sparse salient features, but also to sample matches in non-salient regions that later form two-view-only tracks. However, we found that with the MASt3R~\cite{leroy2024grounding} matcher, using these sampled two-view tracks degrades performance due to low match precision.
\section{Efficiency Analysis}

\begin{table}[t]
\centering
\small %
\begin{tabular}{clcccc}
\toprule
 (seconds)&  \# images: & 25 & 44 & 110 & 505 \\ \hline
\multirow{3}{*}{\rotatebox[origin=c]{90}{\parbox[c]{2cm}{\centering Global \\ Refinement}}} & \makecell{Bundle \\ Adjustment} & 2.76 & 18.8 & 45.4 & 610 \\ 
 & \makecell{\textcolor{blue}{Depth} \\ \textcolor{blue}{Refinement}} & 4.39 & 19.6 & 44.3 & 268 \\ 
 & \makecell{\textcolor{blue}{3D Point} \\ \textcolor{blue}{Covariances}} & 0.96 & 6.49 & 11.5 & 64.4 \\ \hline
\multirow{3}{*}{\rotatebox[origin=c]{90}{\parbox[c]{2cm}{\centering Local \\ Refinement}}} & \makecell{Bundle \\ Adjustment} & 0.47 & 1.91 & 2.34 & 19.1 \\ 
 & \makecell{\textcolor{blue}{Depth} \\ \textcolor{blue}{Refinement}} & 0.21 & 0.73 & 1.19 & 9.94\\ 
 & \makecell{\textcolor{blue}{3D Point} \\ \textcolor{blue}{Covariances}} & 0.80 & 2.57 & 4.66 & 55.5 \\ \hline
\multirow{4}{*}{\rotatebox[origin=c]{90}{\parbox[c]{2.7cm}{\centering Post Registration \\ Refinement}}} & \makecell{3D Point \\ Refinement} & 0.20 & 0.63 & 0.85 & 7.50 \\ 
 & \makecell{\textcolor{blue}{Depth} \\ \textcolor{blue}{Refinement}} & 2.33 & 2.71 & 15.9 & 40.8\\ 
 & \makecell{\textcolor{blue}{3D Point} \\ \textcolor{blue}{Covariances}} & 0.50 & 2.00 & 3.78 & 33.9 \\ 
 & \makecell{\textcolor{blue}{Depth check}} & 0.44 & 0.73 & 2.01 & 13.9 \\ 
 \bottomrule
\end{tabular}
\caption{\textbf{Efficiency analysis of the significant components of our pipeline.} Results, cumulated over all function calls, are presented for reconstructions of scenes from ETH3D \cite{schops2017multi} of sizes 25, 44, 110 and 505 images. The components we add to COLMAP are highlighted in \textcolor{blue}{blue}.
}
\label{tab:efficiency_analysis}
\end{table}

\Cref{tab:efficiency_analysis} presents the efficiency analysis of our pipeline for three scenes with 25, 44, 110 and 505 images. Depth refinement constitutes the most significant overhead compared to COLMAP \cite{schoenberger2016sfm}. During post-registration refinement, the depth map of a newly registered image is refined first. If the image passes the depth consistency check, its depth map undergoes another refinement during local refinement. Notably, this second refinement converges faster as the depth map is already close to the optimization minima, evidenced by the reduced processing time. 

The global refinement is performed periodically during the incremental reconstruction pipeline and refines all depth maps.
Since reconstructions may not change significantly between consecutive global refinements, many depth maps remain near their optimization minima.
To avoid unnecessary computation, we compare the total refinement costs of the previous and current calls.
If the costs differ insignificantly, the refinement is skipped.
In reconstructions with 25, 44, 110 and 505 images, 309, 724, 2186 and 10935 refinements were skipped out of 520, 1071, 2860 and 13566 calls, respectively.

In addition to depth refinement, 3D point covariance computation introduces notable overhead. However, the computational cost of the depth consistency check is negligible.

\begin{figure*}[t]
    \centering
    \resizebox{0.8\textwidth}{!}{%
    \begin{minipage}{\textwidth}
    \setlength{\pheight}{0mm}
    \setlength{\pwidth}{0.005\linewidth}
    \setlength{\iwidth}{0.49\linewidth}
    \setlength{\lwidth}{0.49\linewidth \relax}

    \includegraphics[width=\iwidth]{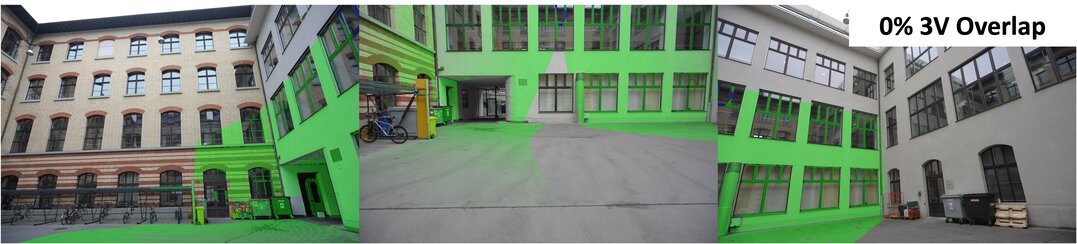}%
    \hspace{\pwidth}%
    \includegraphics[width=\lwidth]{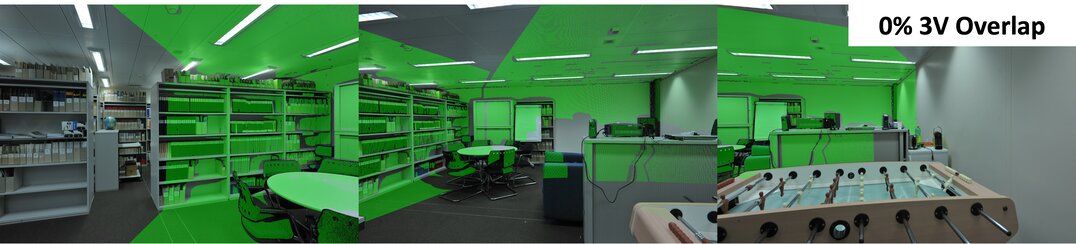}%

    \vspace{\pheight}
    
    \includegraphics[width=\iwidth]{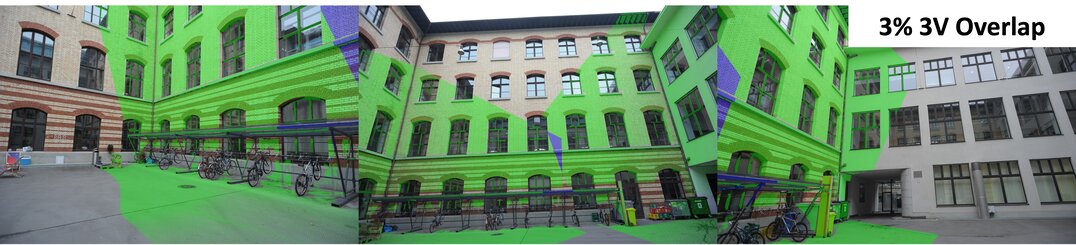}%
    \hspace{\pwidth}%
    \includegraphics[width=\lwidth]{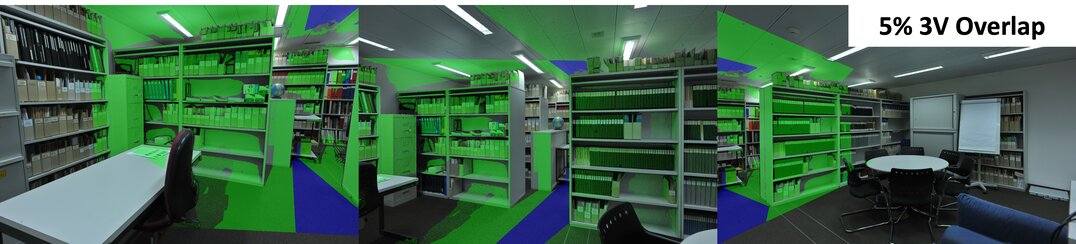}%

    \vspace{\pheight}
    
    \includegraphics[width=\iwidth]{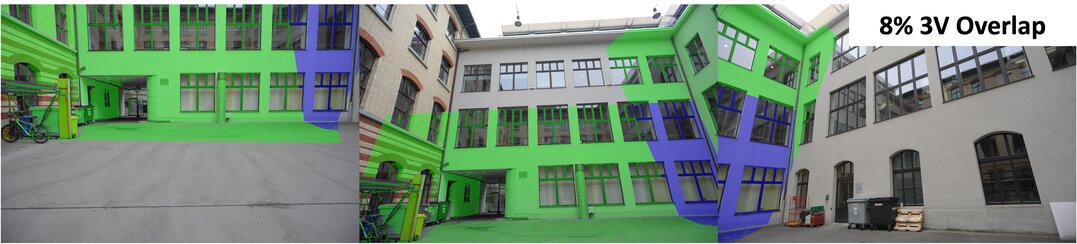}%
    \hspace{\pwidth}%
    \includegraphics[width=\lwidth]{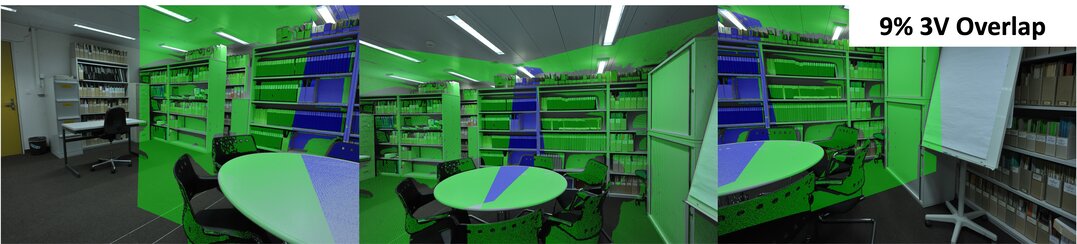}%

    \vspace{\pheight}
    
    \includegraphics[width=\iwidth]{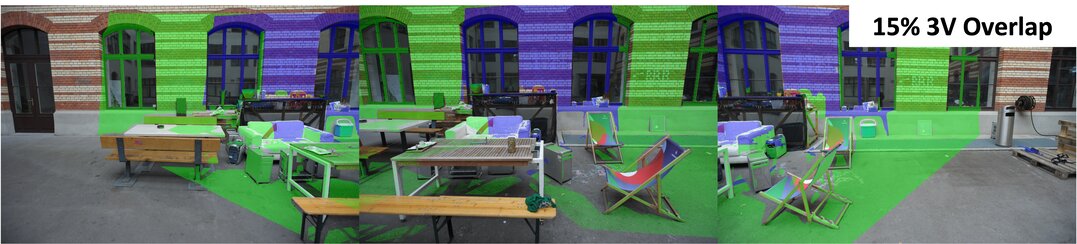}%
    \hspace{\pwidth}%
    \includegraphics[width=\lwidth]{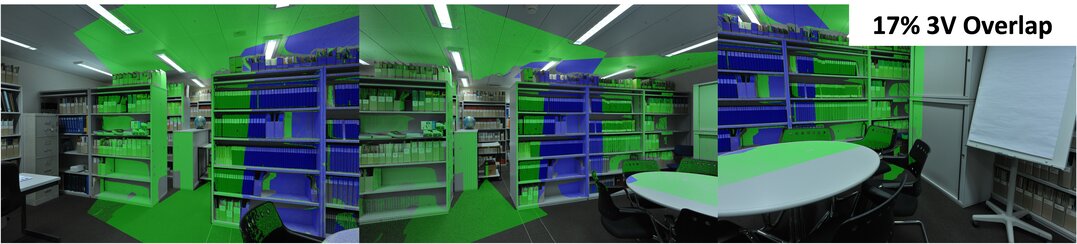}%

    \vspace{\pheight}
    
    \includegraphics[width=\iwidth]{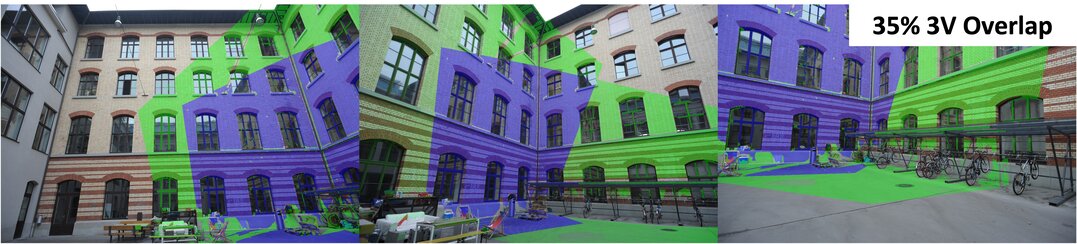}%
    \hspace{\pwidth}%
    \includegraphics[width=\lwidth]{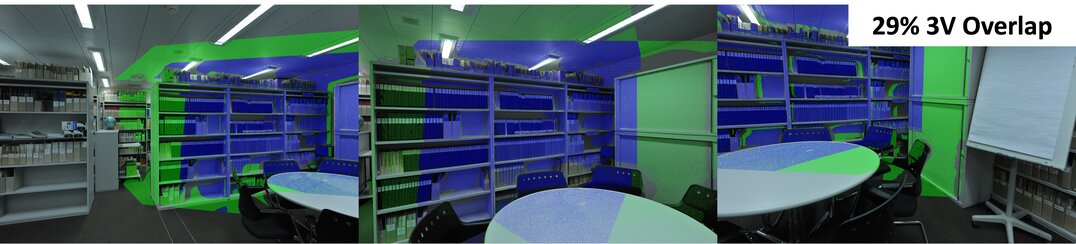}%

    \vspace{\pheight}
    
    \includegraphics[width=\iwidth]{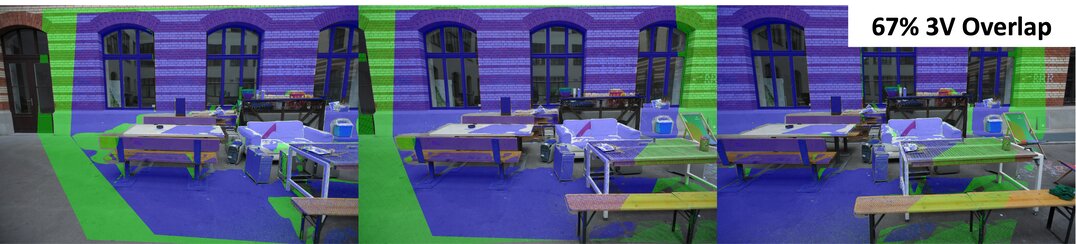}%
    \hspace{\pwidth}%
    \includegraphics[width=\lwidth]{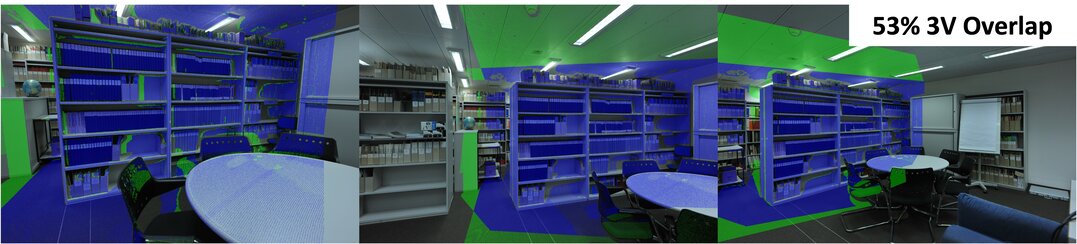}%

    \vspace{\pheight}
    
    \end{minipage}
    }
    \caption{\textbf{Visual examples of our triplet test set.} Each row corresponds to a triplet from one of our triplet test set categories: 0\%, $[0\%,5\%]$, $[5\%,10\%]$, $[10\%,20\%]$, $[20\%,40\%]$, and $>40\%$, respectively. Triplets are colored by {\color[HTML]{33cc33}two-view overlap} and \textcolor[RGB]{31,119,180}{three-view overlap}. All triplets are from the ETH3D \cite{schops2017multi} dataset. Left: examples from the Courtyard scene. Right: examples from the Kicker scene.}
    \label{fig:triplets}%
\end{figure*}

\section{Implementation details}

\subsection{Low-overlap evaluation}

To evaluate the performance of our pipeline under varying levels of image overlap, we constructed test sets by incrementally sampling images based on their overlap with existing images in the scene. This approach allows us to simulate low- to high-overlap scenarios.

\paragraph{Sampling Criteria:}  
Images are added to the test set if they satisfy two conditions:
1) The three-view (3V) overlap with existing test set images does not exceed the target threshold.
2) There is sufficient two-view (2V) overlap with at least one image that has been already selected.

\paragraph{Dataset-Specific Overlap Determination:}  
For each dataset, the overlap is determined using different criteria:
\begin{itemize}
    \item ETH3D \cite{schops2017multi}: for the training set we use the ground truth depth maps; for the test set we use depth maps estimated by multi-view stereo with COLMAP~\cite{schoenberger2016mvs}.
    \item SMERF \cite{duckworth2023smerf} and Tanks and Temples \cite{knapitsch2017tanks}: the overlap is based on the ratio of detected sparse keypoints in an image with the number of common 3D points in a retriangulated SfM point cloud.
\end{itemize}

\paragraph{Minimal test sets:}  
For the minimal (zero 3V overlap) test sets, it is often infeasible to sample images spanning the entire scene. In such cases, images with some 3V overlap are sampled, provided they result in at least one other image with zero 3V overlap.

\paragraph{Test test construction and statistics:}  
For the ``minimal" test sets, we sample 10 sets of images per scene, while all other test sets contain 5 sets per scene. Exceptions exist for ETH3D, where some scenes lack sufficient images to meet the overlap criteria. To ensure diversity,
each image set includes a minimum of 5 images, and 
each image set differs from others within a test set by at least two images.

The average number of images per test set, computed across scenes, is presented below:
\begin{itemize}
    \item ETH3D: \textbf{minimal:} 6.7, \textbf{\textless5\%:} 6.9, \textbf{\textless10\%:} 8.8, \textbf{\textless30\%:} 14.0,\textbf{ all:} 36.9.
    \item SMERF: \textbf{minimal:} 36.4, \textbf{low-overlap:} 78.6, \textbf{medium:} 100.6, \textbf{high:} 170.7.
    \item Tanks and Temples: \textbf{minimal:} 7.8, \textbf{low-overlap:} 17.9, \textbf{medium:} 26.6, \textbf{high:} 45.0.
\end{itemize}

\paragraph{Triplet Test set:}
Visualizations of the triplet test set are presented in \cref{fig:triplets}, ranging from zero to $>40\%$ 3V overlap.

\subsection{Low-parallax evaluation}
For the low-parallax evaluation, we reconstruct scenes from the RealEstate10K dataset \cite{zhou2018stereo}.
Following the sampling strategy of MASt3R-SfM \cite{duisterhof2024mast3r}, we randomly sample 10 images per scene from 1.8k randomly selected videos.

\subsection{Baselines}
For all baselines, except MASt3R-SfM \cite{duisterhof2024mast3r} and VGG-SfM \cite{wang2023vggsfm}, and our approach, we select image pairs using NetVLAD top 20 retrieval. We run both MASt3R-SfM and VGG-SfM with their default approach. 

For COLMAP-based baselines, including COLMAP and its variants (Structureless Resectioning and Detector-free SfM), we use the default hyperparameters provided by COLMAP. Additionally, we fix the intrinsics to their ground truth values in all experiments.

To adapt COLMAP \cite{schoenberger2016sfm} for low-parallax scenarios, we tune its triangulation angle hyperparameters.
Specifically, the minimum triangulation angle is set to 0.001 for both initialization and during the main loop of the incremental mapper.
Additionally, the same minimum triangulation angle of 0.001 is applied during 3D point filtering.

For MASt3R-SfM and VGG-SfM, we similarly use ground truth intrinsics, and fix them during optimization, which yielded the best results. To reconstruct a subset of the high-overlap scenes in ETH3D and Tanks \& Temples, as well as the low-overlap scenes in the SMERF dataset, we reduced the number of keypoints in VGG-SfM. This was necessary to avoid running out of memory, but likely impacts reconstruction quality.

\begin{figure}[t]
    \centering
    \setlength{\pheight}{1mm}
    \setlength{\bwidth}{2.5cm}
    \setlength{\pwidth}{0.005\linewidth}
    \setlength{\iwidth}{\dimexpr(0.999\linewidth - 2\pwidth)/2 \relax}

    \includegraphics[width=0.7\linewidth]{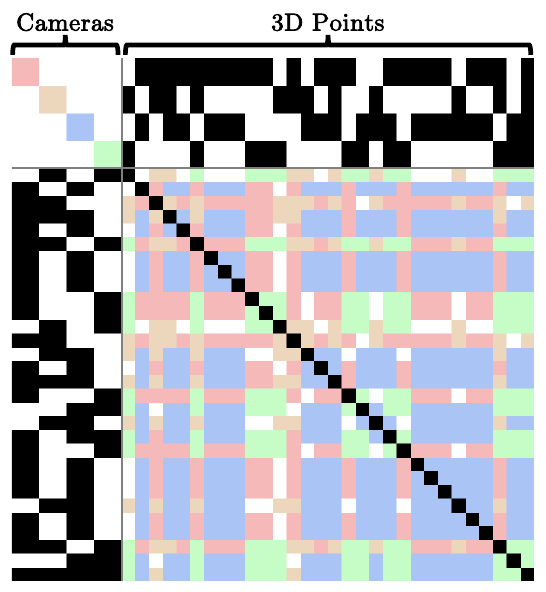}%

    \caption{\textbf{Sparsity structure in the Hessian of the cost function.} Off-diagonals in the 3D point block are color-coded according to the observing camera, reflecting the per-image normal constraints.}
    \label{fig:hessian}%
\end{figure}

\section{Structure of the Hessian}
In \cref{fig:hessian}, we present the Hessian structure of our overall cost function proposed in \cref{sec:local-global-refinement}. Normal and depth constraints couple 3D points and off-diagonal terms in the point block of the Hessian matrix. This breaks the block diagonal assumption required for the Schur complement. 
\section{Additional Ablations}
\label{sec:additional_ablations}

\subsection{Low- to high-overlap dense reconstruction}
\begin{figure*}[t]
    \centering
    \includegraphics[width=\linewidth]{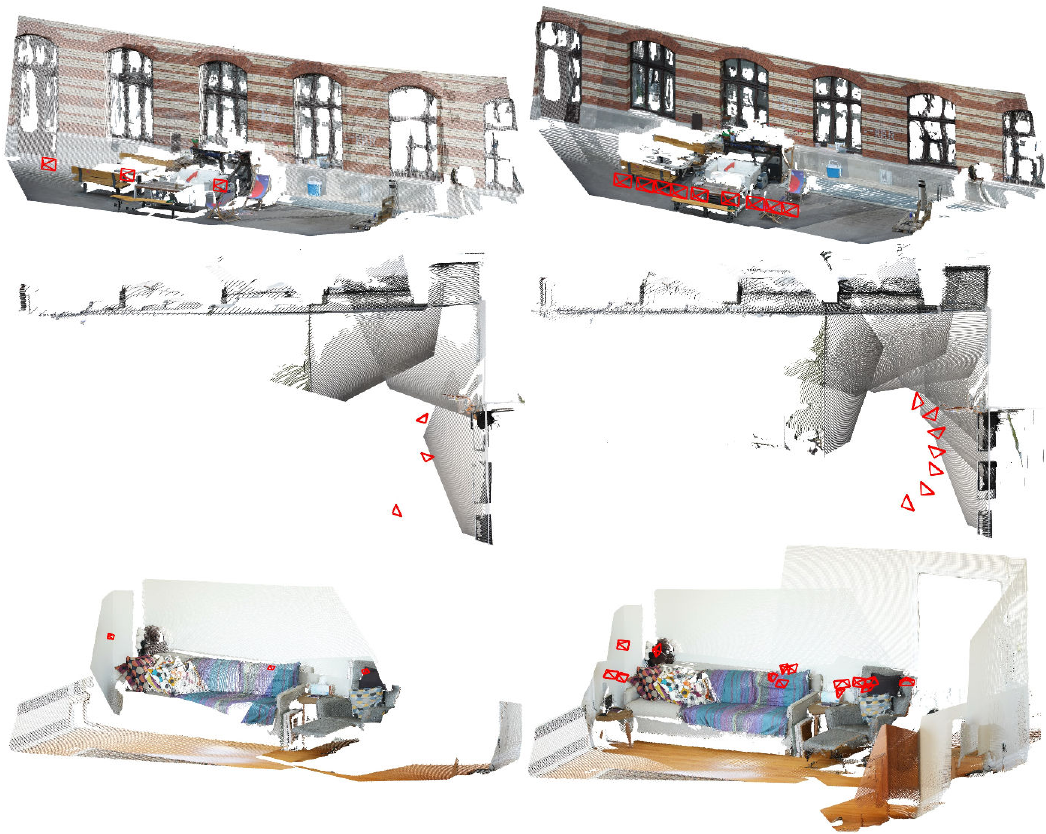} 
    \caption{\textbf{Comparison between low and high overlap dense reconstruction.} Left: Sparse view reconstruction. Right: dense view reconstruction of the same scene. Multiple views constraining the depth refinements yield consistent depth maps. 
    }

    \label{fig:low_high}%
\end{figure*}

In \cref{fig:low_high}, we compare sparse and dense view reconstruction. Our proposed bundle adjustment jointly optimizes camera poses, 3D points, and depth maps. As a result, the point clouds derived from the refined depth maps are well aligned—particularly evident in the second row when observing the building walls.

\subsection{Robust reconstruction despite noisy priors}
\begin{figure*}[!ht]
    \centering
    \includegraphics[width=0.86\linewidth]{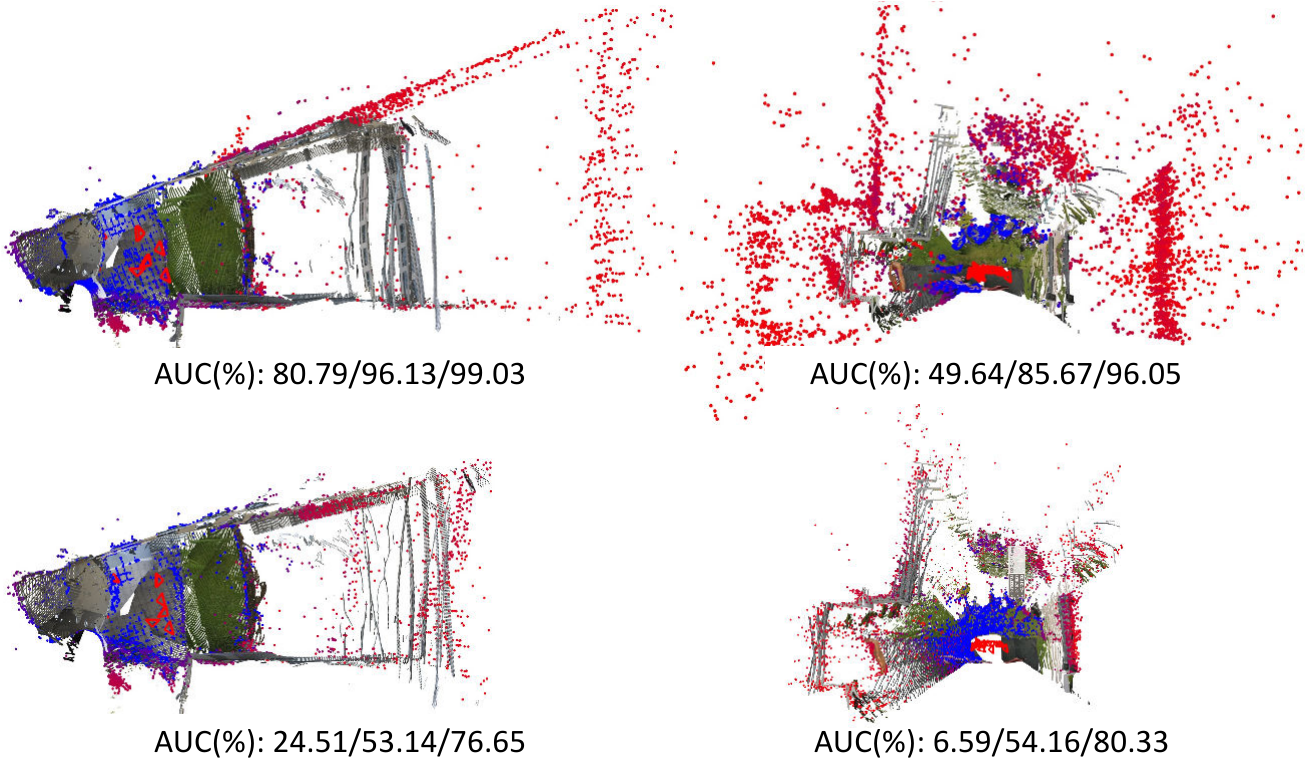} 
    \caption{\textbf{Comparing reconstruction quality with and without robust loss.} Depth Anything V2~\cite{depth_anything_v2} struggles to estimate depth at large distances. We visualize 3D points with {\color{blue}low} and {\color{red}high} covariance, overlaid on the lifted, refined depth maps.
    \textbf{Top:} Reconstruction using a robust loss in the depth term. \textbf{Bottom:} Same scene without the robust loss. Without it, high-covariance points converge to the noisy depth prior, leading to lower reconstruction precision. In contrast, our method achieves accurate reconstruction despite unreliable depth estimates.}

    \label{fig:robust}%
\end{figure*}

To demonstrate the importance of the robust loss in our objective (\cref{sec:local-global-refinement}), we compare reconstructions with and without applying it to the depth term. Despite inaccurate depth priors, our method achieves accurate results. During reconstruction, depth maps are refined and residuals between depths and 3D points are reduced. We observe that using a smaller robust loss scale in the final global bundle adjustment leads to the best accuracy.

\subsection{Next view selection}
\label{sec:next_view_selection}
\begin{table}[t]
\centering
\scriptsize
\setlength\tabcolsep{2pt}%
\renewcommand{\b}[1]{\textbf{#1}}
\begin{tabular}{lcccc}
\toprule
\multirow{2}{*}[-0.4em]{variant} & \multicolumn{2}{c}{ETH3D dataset} & \multicolumn{2}{c}{SMERF dataset}\\
\cmidrule(lr){2-3} \cmidrule(lr){4-5}
& min. overlap & all images & min. overlap & high overlap\\
\midrule
SP+LG + \textbf{ours}
& 27.3/55.9/71.8 & 74.3/88.3/92.0 & 9.2/41.0/69.8 & 47.3/79.3/90.6
\\

num. of vis. points
& 25.4/48.9/62.6 & 71.8/85.2/88.5 & 9.4/40.8/68.0 & 44.2/67.9/76.6
\\
num. inlier corresp.
& 26.7/54.9/70.4 & 74.8/88.6/92.0 & 8.9/40.2/67.6 & 45.8/77.2/90.5
\\

\midrule
ROMA + \textbf{ours}
& 33.4/60.6/74.4 & 71.6/87.0/91.3 & 10.6/41.0/61.8 & 41.4/69.3/79.4
\\

num. vis. points
& 33.5/60.1/73.4 &     68.7/83.8/88.0  & 9.7/35.7/51.2 &    41.1/62.2/70.6
\\
num. inlier corresp.
& 33.5/61.1/75.5 &     70.1/85.5/89.9   &  10.9/41.5/61.9 &    39.3/69.7/81.1 
\\

\midrule
MASt3R + \textbf{ours}
&  34.9/67.2/81.7 &     70.3/88.2/93.6 & 17.2/54.6/77.1 & 56.5/84.4/94.0
\\

num. vis. 3D points
& 32.9/63.7/77.4 &     64.9/81.7/86.7  & 14.2/46.4/65.4 &     51.6/71.9/78.4
\\
num. inlier corresp.
& 33.3/64.0/77.8 &     64.0/80.2/85.2   &  14.1/46.9/68.9 &     54.5/78.8/86.2
\\

\bottomrule
\end{tabular}
\caption{\textbf{Ablation of the next view selection.}
We compare three approaches applicable in our pipeline. In contrast to prior works~\cite{schoenberger2016sfm}, \emph{number of visible 3D points}, includes 3D points with a track length one. \emph{Num. inlier corresp.} selects views by counting the maximum number inlier correspondences between query and registered images. This allows our pipeline to effectively reconstruct low-overlap environments. Counting the sum of feature matcher scores between these image pairs instead leads to the best overall performance.}
\label{tbl:results:ablation_nvs}%
\end{table}

In \cref{tbl:results:ablation_nvs}, we ablate the impact of different next view selection approaches in our pipeline. \emph{COLMAP}’s~\cite{schoenberger2016sfm} next view selection maximizes the robustness of registration in traditional SfM pipelines. A natural adaption of this approach that adheres to our registration method is to count the \emph{number of visible 3D points}, including those with track length one. As a result, however, the next view selection score scales with the number of registered images in local bundles, leading to frequent incorrect selection in the case of symmetries.

To handle all levels of overlap, we, instead rely on two-view information. While selecting the next view with the maximum \emph{number of inlier correspondences} to any registered image would be unstable in traditional SfM, lifting 3D points via monocular depth makes the registration robust. However, the performance of this greedy approach still largely depends on the quality of the matches.

Instead of counting the number of correspondences between query and registered image pairs, we sum their scores (as predicted by the feature matcher). This leads to better performance in our SP+LG-based~\cite{superpoint, lindenberger2023lightglue} pipeline. For dense matchers, selecting next views based on inlier correspondences can cause severe failure cases, especially with \emph{MASt3R}~\cite{leroy2024grounding}, which often hallucinates inlier matches through surfaces. Leveraging matcher scores instead leads to drastic improvements. In the case of \emph{RoMa}~\cite{edstedt2024roma}, the performances of the two approaches are similar.

    \fi
\fi

{
    \small
    \bibliographystyle{ieeenat_fullname}
    \bibliography{abbreviations,main}
}

\ifproceedings
    \ifaddappendix
        \clearpage
        
    \fi
\fi

\end{document}